\documentclass{article}

\usepackage{arxiv}%
\usepackage{graphicx}%
\usepackage{multirow}%
\usepackage{amsmath,amssymb,amsfonts}%
\usepackage{amsthm}%
\usepackage{mathrsfs}%
\usepackage{xcolor}%
\usepackage{textcomp}%
\usepackage{manyfoot}%
\usepackage{booktabs}%
\usepackage{algorithm}%
\usepackage{algorithmicx}%
\usepackage{algpseudocode}%
\usepackage{listings}%

\usepackage{comment}
\usepackage{makecell}
\usepackage{hyperref}
\usepackage{float}
\usepackage{caption}
\usepackage[caption=false,font=normalsize,labelfont=sf,textfont=sf]{subfig}
\usepackage{lineno}

\theoremstyle{thmstyleone}%
\theoremstyle{thmstyletwo}%

\theoremstyle{thmstylethree}%
\newcommand{\set}[1]{\ensuremath \mathbf{#1}} 

\DeclareMathOperator*{\argmin}{arg\,min\,}

\begin{document}

\title{Actionable counterfactual explanations using Bayesian networks and path planning with applications to environmental quality improvement \thanks{This work has been submitted to the IEEE for possible publication. Copyright may be transferred without notice, after which this version may no longer be accessible.

© 2025 IEEE. Personal use of this material is permitted. Permission from IEEE must be obtained for all other uses, including reprinting/republishing this material for advertising or promotional purposes, collecting new collected works for resale or redistribution to servers or lists, or reuse of any copyrighted component of this work in other works.}}


\author{
 Enrique Valero-Leal \\
  Computational Intelligence Group\\
  Departamento de Inteligencia Artificial\\
  Universidad Politécnica de Madrid \\
  \texttt{enrique.valero@upm.es} \\
   \And
 Pedro Larrañaga \\
  Computational Intelligence Group\\
  Departamento de Inteligencia Artificial\\
  Universidad Politécnica de Madrid \\
  \texttt{pedro.larranga@fi.upm.es} \\
  \And
 Concha Bielza \\
  Computational Intelligence Group\\
  Departamento de Inteligencia Artificial\\
  Universidad Politécnica de Madrid \\
  \texttt{mcbielza@fi.upm.es} \\
}

\maketitle

\begin{abstract}
Counterfactual explanations study what should have changed in order to get an alternative result, enabling end-users to understand machine learning mechanisms with counterexamples. Actionability is defined as the ability to transform the original case to be explained into a counterfactual one. We develop a method for actionable counterfactual explanations that, unlike predecessors, does not directly leverage training data. Rather, data is only used to learn a density estimator, creating a search landscape in which to apply path planning algorithms to solve the problem and masking the endogenous data, which can be sensitive or private. We put special focus on estimating the data density using Bayesian networks, demonstrating how their enhanced interpretability is useful in high-stakes scenarios in which fairness is raising concern. Using a synthetic benchmark comprised of 15 datasets, our proposal finds more actionable and simpler counterfactuals than the current state-of-the-art algorithms. We also test our algorithm with a real-world Environmental Protection Agency dataset, facilitating a more efficient and equitable study of policies to improve the quality of life in United States of America counties. Our proposal captures the interaction of variables, ensuring equity in decisions, as policies to improve certain domains of study (air, water quality, etc.) can be detrimental in others. In particular, the sociodemographic domain is often involved, where we find important variables related to the ongoing housing crisis that can potentially have a severe negative impact on communities.
\end{abstract}


\twocolumn

\section{Introduction}

The interest in explainable artificial intelligence \cite{gunning2019xai} (XAI) has grown, as it offers potential solutions to performance, ethical and legal challenges. 
One such type of methods are counterfactual explanations, conditional statements such as ``If it were the case that A, then it would be the case that C'', where A represents a hypothetical past action and C a consequence. In other words, it examines what the outcome would be if a past alternative had occurred. Similarly, we can also study how the past A should have been in order to get an alternative outcome of our desire C. Counterfactuals have recently moved beyond the causality field~\cite{pearl2009causality} towards interpretable machine learning~\cite{guidotti2024counterfactual}, where rather than studying the effects of an external intervention to the current world (Pearl's structural counterfactuals~\cite{pearl2009causality}) they are understood as similar instances with a different trait of interest (typically the class variable).

\begin{figure*}[!t]
    \centering
    \subfloat[]{\includegraphics[width=.42\linewidth]{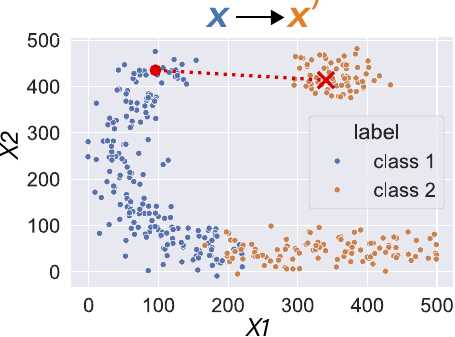}
    \label{fig:overview_a}}
    \hfil
    \subfloat[]{\includegraphics[width=.42\linewidth]{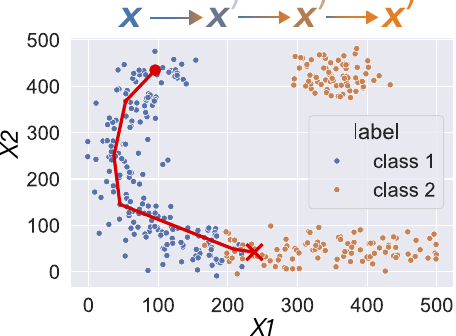}
    \label{fig:overview_b}}

    \subfloat[]{\includegraphics[width=.9\linewidth]{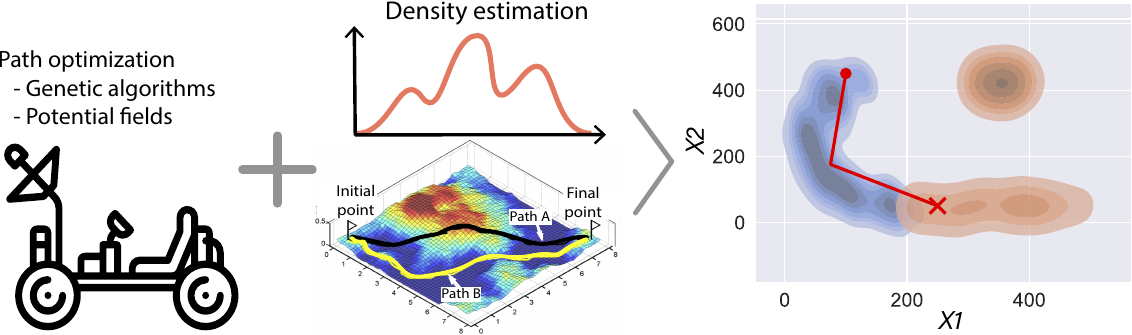}
    \label{fig:overview_c}}

    \subfloat[]{\includegraphics[width=.42\linewidth]{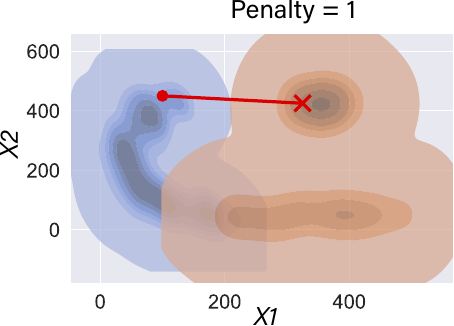}
    \label{fig:overview_d}}
    \hfil
    \subfloat[]{\includegraphics[width=.42\linewidth]{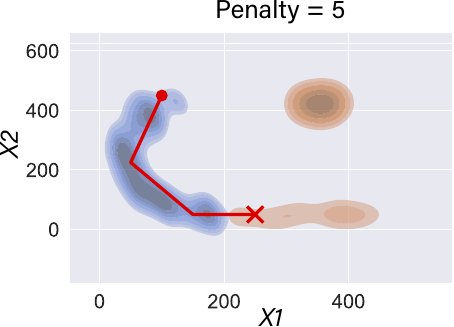}
    \label{fig:overview_e}}

    \caption{Data agnostic actionable counterfactual explanations (DAACE). (a) A traditional approach to find counterfactuals explanations: looking for a similar point $\set{x}'$ to the original instance $\set{x}$ by minimizing the distance. (b) FACE: this work is based on the hypothesis stating that if there is a path of instances between the original instance to be explained (explainee) and the counterfactuals, there exists actionability.  (c) DAACE: we first learn the density of the data and then apply a path planning algorithm using the density as landscape. Endogenous training points are never used as explanations, preserving privacy. (d-e) The parameter penalty controls adherence to the data manifold by penalizing more the low-density areas via a non-linear transform over the negative log-likelihood function.
    }
    \label{fig:overview}
\end{figure*}

There are different metrics to optimize when finding counterfactuals, such as the likelihood of the counterfactual or the Euclidean distance between the instance to be explained and the counterfactual (Fig. \ref{fig:overview_a}). 

Feasible and actionable counterfactual explanations (FACE)~\cite{poyiadzi2020face} focus on finding counterfactuals that users can realistically achieve, proposing to use density-aware algorithms and distance metrics. Such counterfactuals should be connected to the original instance by a path of training points (Fig. \ref{fig:overview_b}).
FACE directly leverages training data and uses an exhaustive search over a graph built on the training data points. This poses significant challenges when scaled to larger datasets or even to fields where access to training data is impossible, which is an emerging issue due to increasing legislation in data privacy.

In contrast, we propose data-agnostic actionable counterfactual explanations (DAACE), an approach based on path planning that aims to overcome the limitations of FACE. Inspired by potential fields~\cite{khatib1986real}, we create a landscape to explore using any specific path planning algorithm~\cite{lavalle2006planning}. The landscape is based on the negative logarithmic likelihood (log-likelihood) function of a trained data density estimator. The instance to be explained is the beginning of the path and each counterfactual is presented as a possible ending point (Fig. \ref{fig:overview_c}). We evaluate the quality of a path by computing its line integral over the negative log-likelihood function~\cite{poyiadzi2020face}. Our algorithm allows for controlling the path's adherence to the data manifold using a penalty parameter (Figs. \ref{fig:overview_d} and \ref{fig:overview_e}).

We particularly focus on Bayesian network-based actionable counterfactual explanations (BayesACE), an instance of DAACE that uses Bayesian networks as density estimators. The use of transparent models instead of black-box models can provide a better potential for explainability~\cite{rudin2019stop}. Bayesian networks explicitly represent the conditional (in)dependencies between variables using a directed acyclic graph (DAG), which can lead to more interpretable and transparent results.

We first experimented using synthetic benchmark datasets to compare both DAACE and BayesACE with FACE, our baseline for actionability. Then, we decide to test our proposals in the active field of environmental quality. BayesACE is applied to policy making for quality of life using the environmental quality index \cite{eqi2020} (EQI) dataset of the Environmental Protection Agency (EPA), as it is important to propose realistic actual policies to apply. We prove the usefulness of BayesACE in supporting policies to improve EQI through actionable counterfactual explanations of different counties in the United States of America (USA).

\section{Background}
\subsection{Probabilistic models and density estimation}
\subsubsection{Normalizing flows}
\label{sec:nf_architecture}
A normalizing flow (NF) \cite{papamakarios2021normalizing} is a generative model that enables the modeling of data distributions through a series of invertible and differentiable transformations \( f = f_1 \circ f_2 \circ \cdots \circ f_T \). Each transformation \( f_i \) maps a simple base probability density function (PDF) \( p_\set{Z}(\set{z}) \) to a more complex PDF \( p_{\set{X}}(\set{x}) \). The change of variables formula allows us to express the density \( p_{\set{X}}(\set{x}) \) in terms of the simpler density \( p_\set{Z}(\set{z}) \) (usually a standardized Gaussian) and the Jacobian determinant of the transformation $f_i$:

\begin{equation*}
p_{\set{X}}(\set{x}) = p_\set{Z}(f_i^{-1}(\set{x})) \left| \det \left( \frac{\partial f_i^{-1}(\set{x})}{\partial \set{x}} \right) \right|.
\end{equation*}

They also allow for the learning of conditional densities:

\begin{equation*}
p_{\set{X} \mid \set{Y}}(\set{x} \mid \set{y}) = p_\set{Z}(f_i^{-1}(\set{x} \mid \set{y})) \left| \det \left( \frac{\partial f_i^{-1}(\set{x} \mid \set{y})}{\partial \set{x}} \right) \right|.
\end{equation*}

We use the well-known RealNVP architecture \cite{dinh2017density}, which uses affine coupling layers to create bijective transformations that are easily invertible and have tractable Jacobian determinants, allowing both efficient sampling and density estimation since both operations will be required in our practical implementation. The input data is split into two subsets in each affine coupling layer. One remains unchanged, while the other is transformed through a learned function, parametrized by neural networks that output scaling and translation transformations.

Thus, for each affine coupling layer, the input vector \( \set{x} \in \mathbb{R}^n \) is split into two disjoint parts, typically \( \set{x}_A \in \mathbb{R}^{\lfloor n/2 \rfloor} \) and \( \set{x}_B \in \mathbb{R}^{\lfloor n/2 \rfloor} \) (or \( \set{x}_B \in \mathbb{R}^{\lfloor n/2 \rfloor +1} \) if $n$ is odd). The outputs of the $l$-th layer, \( \set{x}_{lA} \) and \( \set{x}_{lB} \), are computed as follows:
    
    \begin{align*}
    \set{x}_{lA} &= \set{x}_A, \\
    \set{x}_{lB} &= \set{x}_B \odot \exp(s(\set{x}_A)) + t(\set{x}_A),
\end{align*}

    \noindent where \( s(\cdot) \) and \( t(\cdot) \) refer, respectively, to scale and translation transformations (both affine) and \( \odot \) denotes element-wise multiplication. For greater expressivity, the features are permuted before being input to each layer. This permutation is random, although it has to remain the same for all instances.

    Because \( \set{x}_{lA} = \set{x}_A \), and $s$ and $t$ are functions of only $\set{x}_A$, the inversion is straightforward:
    \begin{align*}
    \set{x}_A &= \set{x}_{lA}, \\
    \set{x}_B &= (\set{x}_{lB} - t(\set{x}_{lA})) \odot \exp(-s(\set{x}_{lA})).
\end{align*}

\subsubsection{Bayesian networks}
A Bayesian network~\cite{koller2009probabilistic} $\mathcal{B} = (\mathcal{G}, \boldsymbol{\theta})$ is a probabilistic graphical model that encodes a joint probability distribution $p(X_1,  ..., X_n)$, by means of only the conditional probability of each node $X_i$ given its parents in a directed acyclic graph. We can model continuous variables alongside discrete ones using conditional linear Gaussian Bayesian networks (CLGNs)~\cite{lauritzen1989clg}, where continuous nodes follow a conditional Gaussian distribution given the values of their parent nodes. Formally, the conditional density $p_{Y \mid \set{X}_d,\set{X}_c}$ of a continuous variable $Y$ given a set of discrete and continuous parents, $\set{X}_d$ and $\set{X}_c$ respectively, is defined as:
    \begin{equation}
    \label{eq:clg}
        p_{Y \mid \set{X}_d,\set{X}_c}(Y \mid \set{X}_d = \set{x}_d, \set{X}_c) \sim \mathcal{N}(b^0_d + \boldsymbol{b}_d^\top \set{x}_c, \sigma_d^2),
    \end{equation}
    \noindent where 
         $\set{x}_d$ and $\set{x}_c$ represent the joint values of discrete and continuous parents, respectively,
         $b^0_d$ and $\boldsymbol{b}_d$ are parameters that depend on $\set{x}_d$, and
         $\sigma_d^2$ is the variance, which is also dependent on $\set{x}_d$.

CLGNs can then model classification problems with numeric features. The structure of the CLGN will follow a Bayesian network-augmented naive Bayes classifier~\cite{bielza2014discrete}, where the target variable is modeled as a discrete root node, with no incoming arcs from the rest of the network nodes, which are continuous and follow a conditional Gaussian distribution (Equation (\ref{eq:clg})). This structure allows for maximum flexibility in the arcs learned while still working exclusively with conditional linear Gaussian distributions.

\subsection{Counterfactual explanations and FACE algorithm}
Let $\set{x} = (x_1,...x_n)$ be an instantiation of variables $\set{X} = (X_1,...,X_n)$ (whose domain $\Omega_{\set{X}} \subseteq \mathbb{R}^n$) labeled $y$, i.e., $(\set{x},y)$. Let $\mathcal{D} = \{(\set{x}^{(i)},y^{(i)})\}^{N}_{i=1}$ be the training data comprised of $N$ instances of a certain classifier, with $\mathcal{D}_{\set{X}} = \{\set{x}^{(i)}\}^{N}_{i=1}$ referring only to the predictor features of the instances within the dataset. We aim to find a counterfactual explanation $(\set{x}',y')$ for the instance $(\set{x},y)$, subject to $y\neq y'$. The idea is to optimize the similarity of $\set{x}'$ to $\set{x}$, rewarding simple explanations but also implicitly showing the shorter way to cross a classifier decision boundary. However, the definition of ``similar'' varies from proposal to proposal, with many of them neglecting if flipping the label to $y'$ is actually possible, i.e., if there is actionability.

The FACE algorithm~\cite{poyiadzi2020face} aims to add actionability to the explanation by hypothesizing that the counterfactual $(\set{x'},y') \in \mathcal{D}$ is actionable if we can find a path (represented by an ordered list) of intermediate instances $\mathcal{P} = \left(\set{x}^{(1)},...,\set{x}^{(v)}\right)$ over the training data ($\set{x}^{(j)} \in \mathcal{D}_{\set{X}}, \ j \in \{1,...,v\}$ and $v \leq N$) from instance $(\set{x},y)$ to counterfactual $(\set{x'},y')$, with $\mathcal{P}$ being the intermediate points between $\set{x}$ and $\set{x}'$. This path captures plausible small changes to the instance that lead to an alternate prediction of its class variable. Theoretically, it proposes using the line integral with respect to a positive transformation of the likelihood function along the path between two consecutive points $\set{x}^{(i)}, \set{x}^{(j)} \in \mathcal{D}_{\set{X}}$ as the metric to minimize, with the selected function typically being the negative logarithm:

\begin{equation}
    \label{eq:face_path_length}
    d(\set{x}^{(i)},\set{x}^{(j)}) = \int_{\gamma} -log(p_{\set{X}}(\gamma(t))) \cdot ||\gamma(t)'||dt,
\end{equation}

\noindent where $p_{\set{X}}(\set{x})$ is the likelihood function (obtained by estimating the density of the data) and $\gamma$ is a bijective parameterization of the straight line that runs between $\set{x}^{(i)}$ and $\set{x}^{(j)}$. The norm of the path differential, $||\gamma(t)'||dt$, represents how we compute the negative log-likelihood for every infinitesimal segment of the path, ``weighting'' its value with the length of said segment.

In practical applications of FACE, the integral is approximated with the likelihood of the middle point between $\set{x}^{(i)}$ and $\set{x}^{(j)}$ or by employing metrics that do not directly use the likelihood function.


\subsection{Path planning}
Path planning involves a set of techniques that, given an initial position \( \set{x} \in \mathbb{R}^n \), a target position \( \set{x}' \in \mathbb{R}^n \) and a set of obstacles \( \mathcal{O} \subset \mathbb{R}^n \), aim to find a continuous mapping \( \mathcal{P} : [0, 1] \rightarrow \mathbb{R}^n \) such that \( \mathcal{P}(0) = \set{x} \), \( \mathcal{P}(1) = \set{x}' \) and $\mathcal{P}(t) \notin \mathcal{O},\ \forall t \in [0, 1]$, representing a path from start to end that avoids obstacles~\cite{choset2005principles}. The path planning goal is further specified by an objective function to optimize particular criteria~\cite{lavalle2006planning}.

One of such techniques upon which our work is based is the theory of potential fields~\cite{khatib1986real}, which models the environment as a field of attractive and repulsive forces acting on the agent, guiding it toward the goal while avoiding obstacles by defining a potential function \( U(\set{x}) : \mathbb{R}^n \rightarrow \mathbb{R} \). This function combines an attractive component \( U_{\text{att}}(\set{x}) \) towards \( \set{x}' \) and a repulsive component \( U_{\text{rep}}(\set{x}) \) against obstacles \( \mathcal{O} \):
\[
U(\set{x}) = U_{\text{att}}(\set{x}) + U_{\text{rep}}(\set{x})
\]

The attractive potential is typically modeled as a function of the distance to the objective, while the repulsive potential discourages proximity to obstacles.

\subsubsection{Non-dominated sorting genetic algorithm II for path planning}
We use the non-dominated sorting genetic algorithm II (NSGA-II)~\cite{deb2000fast} to find optimal paths. This decision is motivated by the efficiency of genetic algorithms, their successful use in path planning~\cite{lamini2018genetic,tuncer2012dynamic}, and the ease of incorporating additional objectives in this particular implementation if needed, as it supports both single- and multi-objective optimization.

Each solution (individual) is represented by an array of size $n(v+1)$. This ensures that the number of genes matches the number of middle points in the polyline ($v$) plus the counterfactual itself, with each point having $n$ dimensions. The instance to be explained does not need to be represented in the array, as it is not optimized. When $v=0$, this corresponds to optimizing straight-line paths between instance $\set{x}$ and counterfactual $\set{x}'$.

\section{Methodology}
\subsection{Theoretical development of DAACE and its computability}
We first theoretically validate DAACE studying its computability and complexity in relation to FACE and formally presenting the framework.

FACE is in P, a complexity class that encompasses problems solvable within time bounds that can be expressed as a polynomial function of the input size. Specifically, the problem has a complexity upper-bounded by $O(N^{3})$, where $N$ is the number of instances in the training set. This is justified by the Dijkstra algorithm having complexity $O(N^2)$ and having to repeat it for each of the $k$ possible counterfactuals. Since in the worst-case scenario, $k = N$ (all the data points are possible counterfactuals), our upper bound is justified.

While FACE models actionability as the ability to find a path between instances, we tackled the problem using the broad framework of path planning in order to find a path that minimizes the integral of the negative log-likelihood over it. This function will take higher values in low-density regions, penalizing these areas as if they were obstacles. DAACE defines the path as a polyline with zero (a straight line) or more vertices between the explainee instance and the counterfactual.

    Theoretically, DAACE aims to solve the problem of actionable counterfactuals in a manner similar to FACE, but considering the entirety of the feature space $\mathbb{R}^n$ instead of only the instances $\set{x}^{(i)} \in \mathcal{D}_{\set{X}}$. 
    


Regarding the complexity of the algorithm, it is possible to prove to which complexity class our proposal belongs by first tessellating the feature space $\mathbb{R}^n$ in $\mathcal{T} = \{ T_i \}_{i\in I}$, so that $\bigcup_{i \in I} T_i = \mathbb{R}^n$ and $T_i \cap T_j = \emptyset , \ \forall i,j \in I$ for $i \neq j$. Now, if we consider the elements of $\mathcal{T}$ as nodes in a shortest-path search, this reduces to a shortest-path problem, solvable in time upper-bounded by $O(|\mathcal{T}|^{3})$, justified in a similar manner as in the previous section.

\subsubsection{DAACE as a path planning problem}

Although the problem is in P assuming a finite tesellation $\mathcal{T}$, it becomes harder with greater cardinality $|\mathcal{T}|$. With an infinitesimal discretization of $\mathbb{R}^n$ (that is, working directly with $\mathbb{R}^n$), $|\mathcal{T}|$ becomes infinite and uncountable. 

We argue that the problem of finding actionable counterfactuals can be solved using path planning, since the concept of minimizing the path length while avoiding obstacles is also present in a distance-aware metric (see Equation (\ref{eq:face_path_length})). The value of the metric decreases for shorter paths $\gamma$ (shorter integration range), but takes greater values for low-density regions (higher value of the negative log-likelihood function, $-log(p_{\set{X}}(\cdot))$). Intuitively, these regions represent the obstacles $\mathcal{O}$ to be avoided.

First, we redefine a path as a polyline (similarly to FACE), since we consider this geometric element to be far simpler than a continuous mapping. Let the path be represented by an ordered list of pairs of distinct elements $\mathcal{P} = \left(\set{x}^{(i)}\right)^v_{i=1}$. Its length $dist_{L}(\mathcal{P})$ is defined as the sum of the distance-aware metric $d$ (see Equation (\ref{eq:face_path_length})) of the lines running between every pair of adjacent points of the path:
    \begin{equation}
    \label{eq:path_length}
    dist_{L}(\mathcal{P}) = \sum_{i=1}^{v-1} d(\set{x}^{(i)},\set{x}^{(i+1)}).
    \end{equation}

To accommodate DAACE to the path planning framework, we can define it as an optimization problem in which the metric to optimize is the actionability. Let $v$ be a natural number, $\set{x}_{(0)}$ be the original instance to be explained and $r : \set{X} \rightarrow \{0,1\}$  be a function such that $r(\set{x}') = 1$ if $\set{x}'$ is a possible counterfactual for $\set{x}$. The optimal counterfactual $\set{x}'^*$ is the one that has the shortest path $\mathcal{P}^* = \left(\set{x}, \set{x}^{(1)*},...,\set{x}^{(v)*},\set{x}'^*\right)$, which is found by solving the following optimization problem:
    \begin{equation}
    \label{eq:daace_formal}
    \left(\set{x}^{(1)*},...,\set{x}^{(v)*},\set{x}'^*\right) = \argmin_{\left(\set{x}^{(1)*},...,\set{x}^{(v)*},\set{x}'^*\right)} dist_{L}\left(\mathcal{P}^*\right), 
    \end{equation}

    \noindent subject to finding a counterfactual $\set{x}'^*$ that meets user-defined log-likelihood and class posterior probability (given the features) thresholds. These thresholds are also present in the FACE algorithm.

\subsection{Models architecture and parameter-tuning}
\subsubsection{Conditional normalizing flow}
\label{sec:nf_fine_tuning}
Since we are interested in the distribution of the data given a class variable $Y$, we will use a conditional version of this NF, in which the transformations are conditioned on a context variable $y$, \( s(\cdot|y) \) and \( t(\cdot|y) \).

All neurons use the ReLU activation function and their weights $\mathcal{W}$ are initialized using the Xavier method \cite{glorot2010understanding} with a multiplier of 1/3, which is motivated to avoid exploding gradients and overflow problems, as the loss function (negative log-likelihood) is not upper-bounded. Formally, each weight $W_i \in \mathcal{W}$ initially

\begin{equation}
    \label{eq:xavier_init}
    W_i \sim \mathcal{N}\left(0, \frac{1}{3}\sqrt{\frac{2}{inp+out}}\right),
\end{equation}

\noindent where $inp$ and $out$ refer to the number of inputs and outputs of a neuron, respectively.

We learn the parameters of the NF by minimizing the negative log-likelihood function by gradient backpropagation using the Adam algorithm. In order to better learn the distribution, we fine-tune the hyperparameters using Bayesian optimization.

We fix the number of batches to 30 for each dataset, instead of using a fixed batch size for each experiment. This decision ensures a roughly equal number of backpropagation steps across datasets. The relatively large batch sizes provide more stable updates, which is desirable, since we are optimizing a potentially sharp log-likelihood function with an unbounded range. Training is conducted over 500 epochs, leading to a maximum of 15000 backpropagation steps.

The Adam optimizer is complemented by a learning rate scheduler that reduces the learning rate by a factor of 0.9 if no improvements are observed over 30 epochs. Furthermore, if performance worsens over 15 consecutive epochs, the model reverts to an optimal state.

If an overflow occurs at the start of training, the standard deviation of the Gaussian used to sample initial weights (see Equation \ref{eq:xavier_init}) is further reduced.

To further avoid potential overflows caused by the optimization of a not bounded function, we decided to apply a small Gaussian noise in each training step to the training data, hence reducing the sharpness of the log-likelihood function and ensuring a smoother convergence.

The values for the learning rate, the L2 penalty, the number of hidden units and hidden layers in the affine coupling transform, the number of transforms $T$ and the standard deviation of the Gaussian noise are selected through Bayesian optimization. We perform a total of 50 optimization iterations, evaluating the log-likelihood using 10-fold cross-validation in each iteration. The total optimized parameters and their range can be visualized in the Supplementary Tables \ref{tab:bayes_opt} and \ref{tab:params_tuned}.

\subsubsection{Bayesian network classifier}
Using a Bayesian network as a density estimator forms the basis of our approach, which we refer to as BayesACE. The decision of using it is justified because it can have a competitive advantage against deep learning models when it comes to explainability, computational speed and simpler fine-tuning. 

Bayesian networks offer explicit graphical representations of the conditional (in)dependencies, making them useful in scenarios where expert evaluation is important~\cite{angelopoulos2022bayesian}. In addition, they can potentially encode causality.

Both the structure and parameters of the Bayesian network are learned from the data. For the structure, we used a score-and-search approach, where a space of directed acyclich graphs as Bayesian network structures is explored iteratively and the candidate solutions are evaluated based on a scoring function, as the Bayesian information criterion (BIC)~\cite{schwarz1978estimating}. The BIC is a widely used metric that balances model fit and complexity by penalizing overly complex structures, thereby reducing the risk of overfitting and that tends to recover original structures when working with sufficiently large datasets~\cite{kitson2023survey}. Regarding the parameters, they were learned using maximum likelihood estimation, thus maximizing the likelihood of the data given the parameters.

For the experiments regarding the EQI dataset, we imposed additional restrictions in the structure learning procedure to ensure simplicity and interpretability by:

\begin{enumerate}
    \item Forcing an arc from the total EQI to all 5 individual indices,
    \item Restricting arcs from the total EQI to any variable,
    \item Restricting arcs flowing from a domain-specific index to a variable concerning a different domain,
    \item Restricting arcs between variables of different domain and
    \item forbidding variables to have more than 3 parents in the graph, in order to ensure simplicity and interpretability.
\end{enumerate}

These restrictions ensure a Bayesian network structure similar to how the quality indices are computed, with the following (in)dependencies holding:
\begin{enumerate}
    \item The overall EQI depends on the domain-specific indices.
    \item The EQI is conditionally independent of the variables of one domain given its domain-specific index.
    \item The domain variables of a certain domain are conditionally independent of the domain variables of a different domain, given any of the corresponding domain-specific indices.
\end{enumerate}

The mean log-likelihood of the data given the Bayesian network is -137.89, whereas removing restriction number 5 (concerning node arcs) increases to -123.35. For reference, the mean log-likelihood of the data given RealNVP is -98.93.

\subsubsection{NSGA-II to solve DAACE}
Two restrictions are incorporated to NSGA-II: a log-likelihood threshold $\alpha \in \mathbb{R}$ and a class posterior probability threshold $\beta \in [0,1]$. Given a counterfactual $\set{x}'$ and a user-defined counterfactual class $y'$, then $-log(p_{\set{X}}(\set{x'})) > \alpha$ and $p_{Y|\set{X}}(y'|\set{x}') > \beta$ should hold, where $p_{\set{X}}$ refers to the likelihood of the data (estimated using the NF or CLGN) and $p_{Y|\set{X}}(Y \mid \set{X})$ refers to the probability of the (discrete) class variable given the features. These two restrictions are also present in the FACE algorithm and ensure that the likelihood and confidence (class posterior probability) metrics are not neglected.

The fitness function is the actionability metric, i.e., the integral of the negative log-likelihood of the path (Equation (\ref{eq:path_length})) encoded in each individual.

The initial population is obtained as follows:
\begin{enumerate}
    \item A set of potential counterfactuals $\{\set{x}'^{(j)}\}_{j\in J}$ that comply with the above restrictions are sampled.
    \item If the user sets $v>0$, then we find the $v$ intermediate points between $\set{x}$ and $\set{x}'^{(j)}, \ \forall j \in J$.
    \item Encode each counterfactual $\set{x}'^{(j)}$ and its middle points (if present) on a chromosome.
\end{enumerate}

The initial population will evolve for a user-defined number of generations, optimizing Equation (\ref{eq:daace_formal}) after the process.

We fix the population size to 100 and the number of generations to 1000, although an early stop is allowed if there are no significant improvements in the last 20 generations.

For genetic operators, we used a simulated binary crossover \cite{deb1995sbx} with a crossover probability of 0.5. This operator simulates a multimodal PDF using both parents as the modes and sample 2 offspring from it. We also used polynomial mutation \cite{deb2008mutation} with a probability of 0.9, which models a PDF from the original individual (the mode is centered on its original values) and modifies its genes sampling from it.

The dispersion from the modes of the aforementioned PDFs is controlled by a dispersion parameter $\eta$, with low values yielding a distribution with a higher dispersion. The value of $\eta$ for both the crossover and mutation operators is selected for each dataset by optimizing for the best counterfactuals in a set of instances using grid search. The values for the crossover $\eta$ range between 10 and 20, while the mutation has a much larger range of 10 and 90. The selection operator (either tournament or random selection) is chosen using the same grid search.

\section{Experiment design}
\label{sec:exp_design}
\subsection{Datasets}
We use a total of 15 synthetic non-normal datasets, as well as a simple 2D toy dataset to illustrate the proposal (see Fig. \ref{fig:overview}). The 15 datasets are resampled version of the numerical classification datasets from \cite{grinsztajn2022tree} and details on preprocessing and generation are present in the Supplementary materials.
 
We also work with the EQI dataset, developed by the USA EPA to assess environmental quality in the 3144 USA counties (3142 after removing two counties with missing values). It integrates data from five environmental domains: air, water, land, built (or infrastructure) environment, and sociodemographics. Each domain includes numerous indicators collected from federal, state and local sources between 2006 and 2010, being this the latest collection of such data released by the EPA. The datasets have been recently used in different public health, risk assessment and exploratory studies~\cite{Jagai2021, Wiese2023}.

A domain-specific quality index per variable domain is built by performing principal component analysis (PCA) on the respective domain variables and selecting the first principal component as the domain-specific quality index. Hence, each index depends exclusively on the variables in its domain. Similarly, an overall index (the EQI) is obtained applying PCA to the five domain-specific indices and selecting the first principal component (Fig. \ref{fig:eqi-over_a}). This hierarchical structure for index computation will be considered when learning the structure of the Bayesian network classifier (Fig. \ref{fig:eqi-over_b}).

Concerning the variables, the same preprocessing as that applied to the synthetic data is carried out, which results in the deletion of 13 features from a total of 139. Among them, 11 features belonged to the water domain, where we find a considerable number of variables with sparse unique values.

In the original dataset a high value for an EQI refers to a bad environmental quality. However, throughout this work we prefer to invert this definition to have a more intuitive notion of the EQI: A high index represents high environmental quality.

\subsection{Algorithm implementations used for the experiments}
We learn a conditional NF (RealNVP architecture) and a CLGN on the datasets resampled with kernel density estimation (Supplementary Table \ref{tab:datasets}) to assess and optimize the value of $dist_{L}$ (Equation (\ref{eq:path_length})) of the paths. The aggregated performance of each density estimator can be visualized in Fig. \ref{fig_ext:model_perf}. 

These are the specific implementations of algorithms used in our experiments:

\begin{enumerate}
    \item DAACE GT, which uses the true distribution (i.e., the kernel density estimator used for data resampling) rather than a learned one by a Bayesian network or a NF. It will serve as an upper bound of the DAACE algorithm.
    \item The DAACE algorithm, using RealNVP as density estimator (referred to simply as DAACE).
    \item The BayesACE algorithm, identical to DAACE but using the CLGN to estimate the density. For all DAACE cases, a genetic algorithm is used to optimize Equation (\ref{eq:daace_formal}) and each genetic algorithm will be fine-tuned separately for each model and for each dataset.
    \item FACE GT. This specific implementation of FACE does not use any of the metrics presented in the original work~\cite{poyiadzi2020face} nor any of the learned density estimators. Rather, we use the more theoretical definition of path length presented in the FACE work (Equation (\ref{eq:face_path_length})) and the ground truth distribution (similarly to DAACE GT). This will serve as a theoretical upper bound for FACE and will be our main baseline comparison.
    \item The FACE algorithm using a density estimation-based graph (FACE-DE) with RealNVP as the density estimator. The distance of a graph edge is the negative log-likelihood of the middle point of the edge.
    \item The FACE algorithm with an $\epsilon$-graph (FACE-$\epsilon$), where no density estimator is needed to weight the edges. However, the NF is used to calculate the logarithmic likelihood and class posterior probability thresholds.
    \item Wachter's counterfactual explanations algorithm~\cite{wachter2017counterfactual}, which will provide a notion on how actionable are the algorithms compared to non-actionable ones.
\end{enumerate}

\begin{figure}[!t]
    \centering
    
    \includegraphics[width=\linewidth]{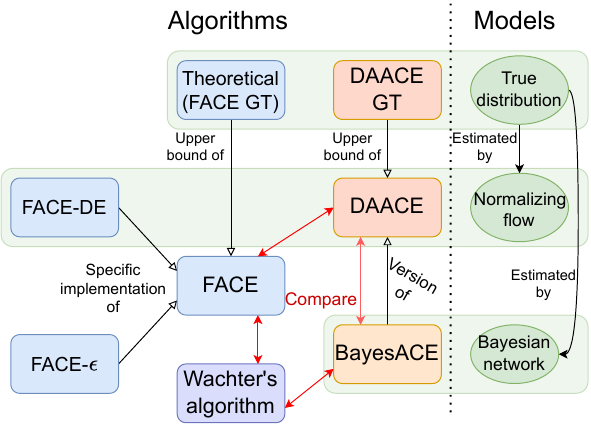}

    \caption{Experiment design overview: we compare a total of seven algorithms, detailed in Section \ref{sec:exp_design}.}
    \label{fig:synt-over}
\end{figure}

The number of nodes in the FACE graphs is 1000, obtained by subsampling at random from the original datasets. This size is reasonable given the number of features considered, which ranges from 4 to 54 (see Supplementary Table \ref{tab:datasets}).

To ensure a fair comparison with DAACE, we set FACE’s connectivity parameter $\epsilon$ to infinity, making the graph fully connected. Instead, we rely on the penalty parameter to regulate the edge weights. Increasing the penalty has the same intuitive effect as reducing the connectivity threshold, as edges that run across low-density areas will receive a greater weight.

Since the number of vertices must be specified in the case of DAACE, we will repeat the experiments with 0, 1, 2 and 3 vertices and the analysis will be done using the algorithm instance that better minimizes actionability (Equation (\ref{eq:daace_formal})).

\subsection{Algorithm parameters and number of experiments}
Since we are dealing with different datasets with very different distributions, we were particularly careful with the log-likelihood and class posterior probability thresholds, $\alpha$ and $\beta$, as they can vary significantly between datasets. Instead of fixing a number on the experiments, we use a multiplier $m_i \in \mathbb{R}$ and, for each dataset, the thresholds $t_i$ for each dataset $i$ will be calculated as follows:

\[ t_i = \mu_{GT_i} + m_i\cdot\sigma_{GT_i}, \]

\noindent where $\mu_{GT_i}$ refers to the mean log-likelihood (for $\alpha$) or mean squared error in predicted probability (for $\beta$) for dataset $i$ and $\sigma_{GT_i}$ to its respective standard deviation.

For each dataset, we find 15 counterfactuals for each different combinations of the penalty parameter (values 1, 5, 10 and 15), log-likelihood threshold multiplier (-1, -0.5 and 0) and class posterior probability multiplier (-0.5 and 0). The combinatorial of parameter values results in a total of 360 counterfactuals per algorithm and per dataset and a total of 5400 counterfactuals per algorithm across all datasets.

The value of the counterfactual class is determined automatically. The class variable is always binary.

\subsection{Standardization of results and statistical testing of hypothesis}
Actionability (density-aware distance) will be calculated using the ground-truth density for all counterfactuals, regardless of the algorithm used to find it. This ensures a higher consistency in the comparisons.

The results for the different datasets vary considerably, given their different nature. We decide to standardize the results for joint interpretation and visualization (Figs. \ref{fig:synt-exp} and \ref{fig:synt-exp-l2}). For each of the 5400 experiments (for each dataset, counterfactual and combination of parameters) we standardize the results using the mean and standard deviation of the results yielded by the seven algorithms considered. As such, all the results shown are relative, which means that algorithms with medians and interquartile ranges below 0 perform on average better (less distance or resources used).

The statistical significance across the algorithms will be measured using the Friedman test \cite{demvsar2006statistical, friedman1937test} adjusted using the Bergmann-Hommel \textit{post hoc} procedure \cite{bergmann1988improvements}, unless explicitly stated. Details on hypothesis testing and its visualization are presented in the Supplementary materials.

\subsection{Experiment with the EQI dataset}
For experiments using the EQI dataset, we slightly adjusted the algorithms and parameter ranges presented above.

First, we limit the number of algorithms used. The ground-truth algorithms are removed since we do not have the ground-truth distribution of the real data. FACE-$\epsilon$ is removed because it is not competitive enough in most metrics in synthetic experiments, except for speed. Finally, since interpretability is key in this experiment and we find BayesACE to be more transparent than DAACE (using a NF), we remove DAACE as well. This leaves us with just BayesACE, FACE and the Wachter's algorithm.

Instead of finding 15 counterfactuals, we find a total of 150 counterfactuals per parameter combination. We experiment with penalties in the range 3, 5, 10 and 15. In this experiment, we fixed the value of the log-likelihood threshold to -125.47, a number obtained by estimating the logarithmic probability of the data given the NF and subtracting 0.25 times its standard deviation. The posterior probability of the class (threshold $\beta$) is fixed to 0.7, which we considered reasonable given that we are dealing with a problem with 7 different class values.

The value of the counterfactual class is set as follows: since we want to improve the quality of life of the counties and we have 7 ordinal classes, we will set the counterfactual class to take the value of the original instance minus 2 (i.e., improvement in two categories the county quality of life).

Since we do not have a ground-truth distribution to compute the true actionability for every algorithm, we decide to use the learn NF flow, as it has a lower negative log-likelihood than the Bayesian network (Supplementary Fig. \hyperref[fig_ext:model_perf]{1b}).

\section{Results}
\subsection{DAACE performance assessed in synthetic datasets}
A summary of the experiments using synthetic benchmark datasets is shown in Fig. \ref{fig:synt-over}, whereas the model performance is shown in Supplementary Fig. \hyperref[fig_ext:model_perf]{1a}. We compare different instances of FACE against DAACE and BayesACE. For DAACE (and also for FACE-DE, an instance of FACE), a NF is used to estimate the data density. We will focus on studying performance across different metrics. The results obtained are standardized and therefore usually range between -3 and 3. Hence, for instance, a negative median distance for an algorithm means that its distance is lower than the average for all algorithms, rather than it's real value being negative.

\subsubsection{Only a middle point is needed for an actionable path}
The shortest feasible actionable recourse path for both DAACE and BayesACE can be found with only two lines, performing significantly better in terms of actionability than when using a different number of lines.

A higher penalty parameter (higher adherence to the data manifold) results in the need for more middle points for DAACE from 1 to 2. For BayesACE, a single middle point is still preferred for higher penalties. In contrast, when the restriction is softened (penalty of 1), both DAACE and BayesACE only require a straight line to represent the actionable recourse.

\begin{figure*}[!t]
    \centering
    \subfloat[]{\includegraphics[width=.94\linewidth]{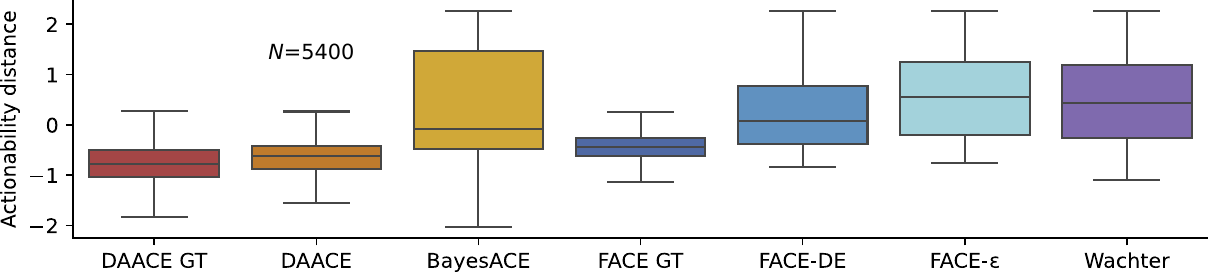}%
    \label{fig:synt-exp_a}}
    
    \subfloat[]{\includegraphics[width=.45\linewidth]{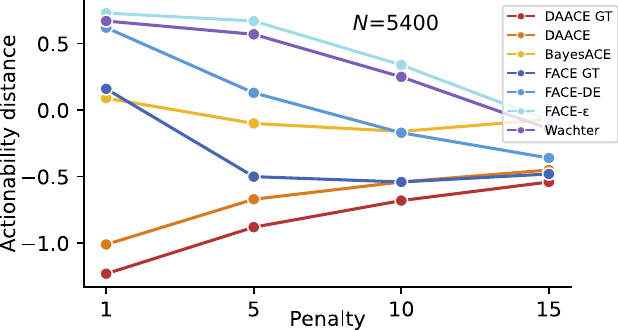}%
    \label{fig:synt-exp_b}}
    \hfil
    \subfloat[]{\includegraphics[width=.45\linewidth]{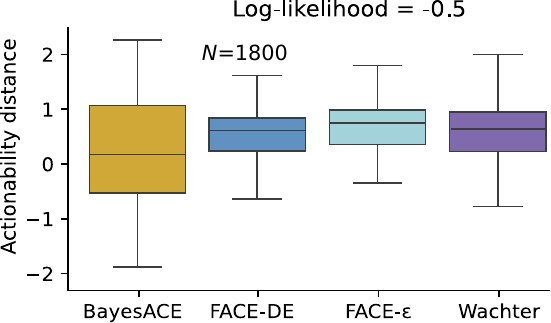}%
    \label{fig:synt-exp_c}}
    \hfil

    \caption{Results using a synthetic benchmark. \textbf{a}, Box plot of the actionability distance, defined as the integral of the negative log-likelihood over the counterfactual path (low distance marks higher actionability): DAACE is significantly more actionable than previous algorithms. BayesACE is competitive against FACE-DE (median of -0.07 against 0.08), even if the difference is not statistically significant (number of samples $N=5400$, for statistical tests and p-value thresholds for null hypothesis rejection refer to the Methods section). \textbf{b}, Line plot of the actionability distance given the penalty: while DAACE and BayesACE performance drops slightly for higher penalties, DAACE GT and DAACE still significantly surpass FACE GT and FACE-DE for all penalties, respectively ($N=5400$). \textbf{c}, Box plot setting a penalty of 1 and a log-likelihood threshold multiplier of -0.5 (a relaxed penalty and log-likelihood threshold). BayesACE performs significantly better than FACE-DE ($N=1800$). Statistical testing of significant differences is shown in Supplementary Fig. \ref{fig_supp:synth_res}.}
    
    \label{fig:synt-exp}
\end{figure*}

\subsubsection{DAACE finds more actionable counterfactuals than FACE}
We find statistically significant differences between algorithms with respect to actionability (Fig. \ref{fig:synt-exp_a} and Supplementary Fig. \hyperref[fig_supp:synth_res]{1a}). In general, DAACE consistently finds shorter paths than FACE, even when comparing FACE GT with DAACE, which uses a learned NF instead of the true distribution (see Supplementary Fig. \hyperref[fig_ext:model_perf]{1a} for a comparison between ground-truth and NF), suggesting an overall more efficient path selection mechanism in DAACE.

BayesACE performs better than FACE-DE in scenarios where we select a low penalty, whereas FACE-DE finds more actionable paths when we set higher penalties (Fig. \ref{fig:synt-exp_b}) or higher likelihood thresholds (Supplementary Fig. \ref{fig_ext:bayesace_logl}). This trend is also noticeable compared to the Wachter algorithm for a high log-likelihood threshold. When BayesACE can find counterfactuals, they usually exceed FACE in actionability (Fig. \ref{fig:synt-exp_c}).

BayesACE, FACE-DE and Wachter's algorithm outperform FACE-$\epsilon$ for almost all parameter combinations. This suggests that the use of a density estimator is necessary.

\subsubsection{BayesACE is more resource efficient, while DAACE is at worst competitive}
The running time to find counterfactuals is significantly longer for DAACE and DAACE GT than for the rest of the algorithms. However, if we take into account the time needed to build the graph over which FACE applies the Dijkstra algorithm DAACE becomes faster than FACE-DE and FACE GT (Supplementary Fig. \ref{fig_supp:synth_res}). Assuming the graph is not always stored in memory is reasonable, as the space in the computer memory scales in order $O(N^2)$, where $N$ refers to the number of endogenous data points in the FACE graph. With a graph of size 100000 this would rise close to 80 gigabytes.

\begin{figure}[!h]
    \centering
    \includegraphics[width=1\linewidth]{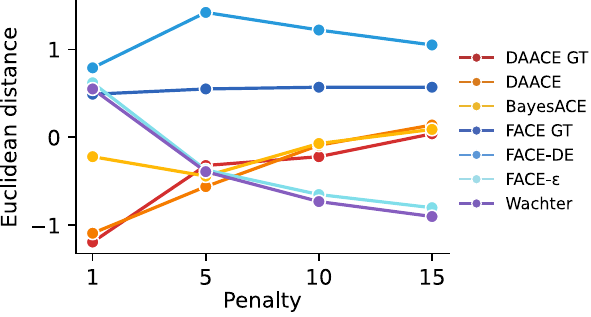}
    \caption{Line plot of the Euclidean distance: DAACE and BayesACE are significantly better than FACE GT and FACE-DE respectively. For lower penalties, DAACE and BayesACE are also significantly better (or competitive) than Wachter's, a non-actionable algorithm ($N=5400$).}
    \label{fig:euclidean_line}
\end{figure}

\subsubsection{BayesACE and DAACE find more similar and sparser counterfactuals}
DAACE consistently finds counterfactuals that are significantly closer in terms of Euclidean distance to the original instance compared to both FACE GT and FACE-DE, even performing better than Wachter's algorithm in this metric for lower penalties (Fig. \ref{fig:euclidean_line}). 

Regarding BayesACE, it generally finds counterfactuals further from the instance than Wachter's and DAACE, but it performs significantly better than FACE GT (only for lower penalties) and FACE-DE, which actually achieves the worst results in this metric. BayesACE can also match Wachter's algorithm for lower penalties and log-likelihood thresholds, as shown in Supplementary Fig. \ref{fig_ext:bayesace_l2}.

\begin{figure*}[!t]
    \centering
    \subfloat[]{\includegraphics[width=.45\linewidth]{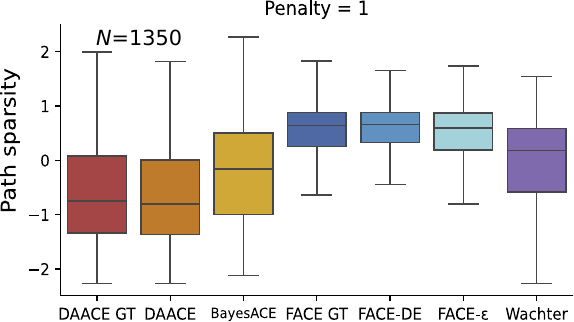}
    \label{fig:synt-exp-l2_a}}
    \hfil
    \subfloat[]{\includegraphics[width=.45\linewidth]{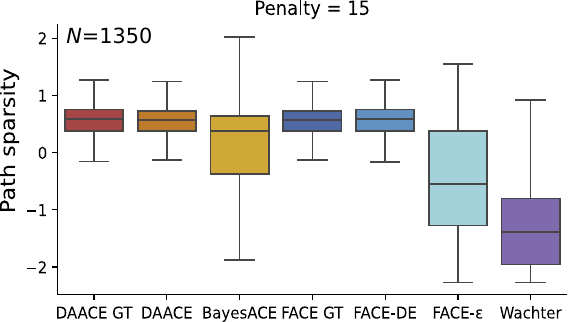}
    \label{fig:synt-exp-l2_b}}

    \subfloat[]{\includegraphics[width=.94\linewidth]{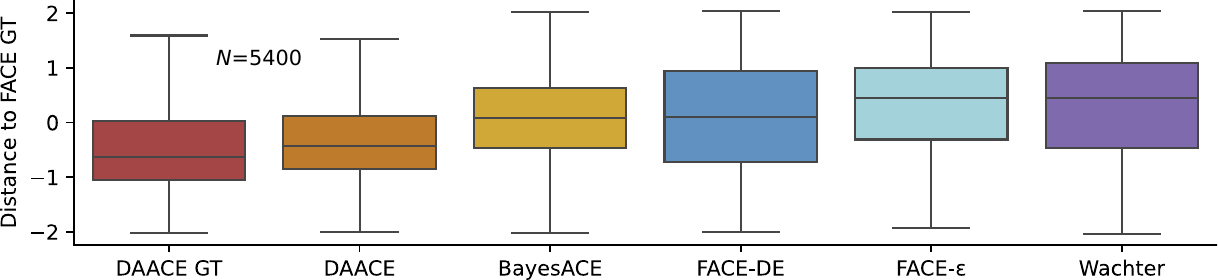}
    \label{fig:synt-exp-l2_c}}

    \caption{Results using a synthetic benchmark. (a,b) Box plots of the path sparsity for different penalties: DAACE performs better for lower penalties, requiring fewer feature changes than the FACE algorithms (except for FACE-$\epsilon$ for higher penalties) and even than Wachter's, which achieve similar levels of sparsity. The median sparsity of the BayesACE paths is still lower (-0.16) and significantly surpasses FACE GT and FACE-DE (medians of 0.65 and 0.64), but the statistical tests still reveal Wachter's to be sparser than BayesACE. For higher penalty values, DAACE nor BayesACE are no longer able to compete with the Wachter's algorithm. Even if the statistical tests reveal FACE GT to be significantly sparser than DAACE with this penalty, their median path sparsity is equal, 0.57. DAACE, BayesACE and FACE-DE are competitive with one another ($N=1350$ for both experiments). (c) Box plot of the difference with the baseline (Euclidean distance to FACE GT): DAACE (median of -0.43) is significantly closer to the baseline than FACE-DE and FACE-$\epsilon$ (medians of 0.1 and 0.45, respectively). BayesACE median proximity (0.08) is also lower than the ones of the FACE implementations, even if we cannot prove that is statistically significantly better ($N=5400$). Statistical testing of significant differences is shown in Supplementary Fig. \ref{fig_supp:synt-l2}.}
    
    \label{fig:synt-exp-l2}
\end{figure*}

We also analyzed the sparsity of the obtained path, i.e., the number of features that changed throughout the path. Wachter's algorithm and FACE-$\epsilon$ modify the lowest number of features per step compared to the rest of the algorithms, although for lower penalties, DAACE and BayesACE are the superior choices (Figs. \ref{fig:synt-exp-l2_a} and \ref{fig:synt-exp-l2_b}).

\subsubsection{DAACE and BayesACE get results closer to the FACE ideal}
We also decided to compare the similarity (Euclidean distance) between the counterfactuals obtained by all the methods and those obtained by FACE GT, our baseline method. DAACE has the potential to replicate the counterfactuals obtained by FACE GT, since the distance between the counterfactuals obtained between FACE GT and DAACE GT is the shortest among all algorithms (Fig. \ref{fig:synt-exp-l2_c}).  BayesACE has the potential to produce more similar results than FACE-DE when the penalty is high (values of 5, 10 and 15) and the log-likelihood multiplier is lower than 0.

\subsection{EQI improvement using BayesACE}
\begin{figure*}[!t]
    \centering
    \subfloat[]{\includegraphics[width=.96\linewidth]{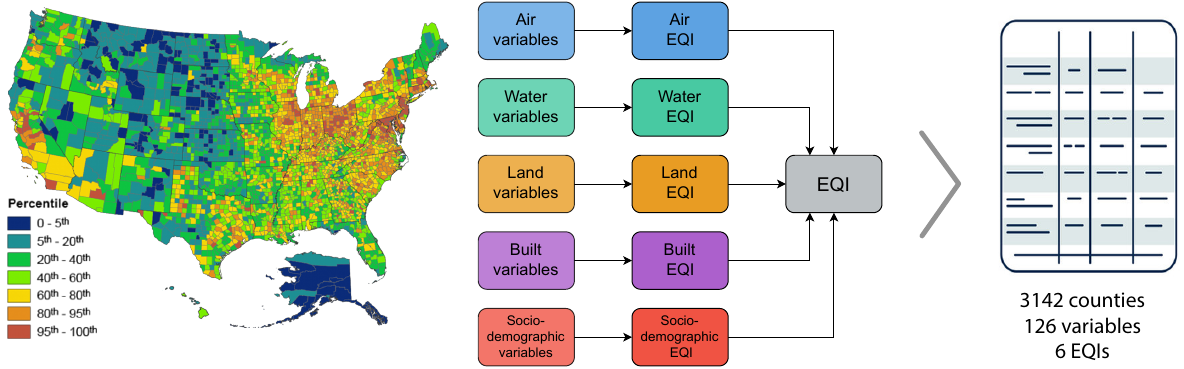}
    \label{fig:eqi-over_a}}

    \subfloat[]{\includegraphics[width=.5\linewidth]{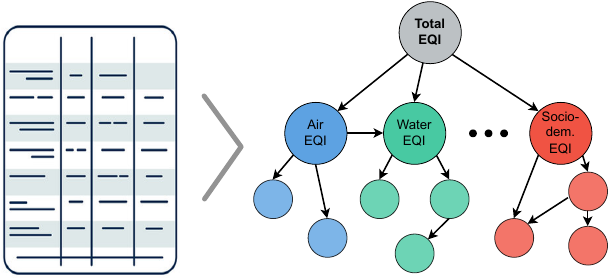}
    \label{fig:eqi-over_b}}
    \hspace{1cm}
    \subfloat[]{\includegraphics[width=.4\linewidth]{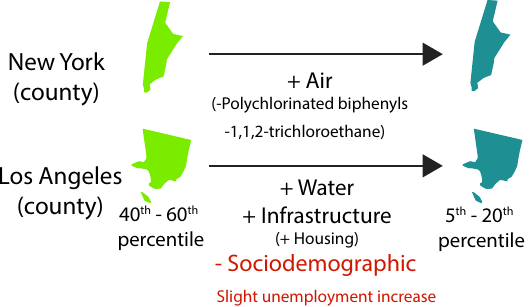}
    \label{fig:eqi-over_c}}
    
    \caption{EQI dataset overview. (a) The map represent the EQI of each USA county aggregated by percentiles (lower means higher quality). The variables are organized into 5 different domains, from where 5 different domain-specific EQIs are derived. Combining them yields the total EQI of a county. We then aggregate the data to learn a model. (b) The structure of the Bayesian network classifier  learned from the data. An instance of Conditional RealNVP is also learned, serving as a better estimation of the ground truth. (c) The BayesACE counterfactuals can be interpreted as a way to improve the counties EQIs in an actionable manner, moving them from the green to light blue category. In this examples, the proposals could have a backlash in the sociodemographic domain.}
    \label{fig:eqi-over}
\end{figure*}

Our proposal can find actionable counterfactuals in the environmental and social domain to improve the quality of life of USA counties (Fig. \ref{fig:eqi-over_a}). In this scenario, we do not have a ground truth. We estimate the data density using both a NF and a Bayesian network classifier (Fig. \ref{fig:eqi-over_b} and Supplementary Fig. \hyperref[fig_ext:model_perf]{1b}). We consider three models: (1) BayesACE to prove the usefulness of Bayesian network classifier and to reassert it as a simple yet effective alternative, (2) the FACE algorithm using a NF to weight the graph edges, which serves as our baseline comparison in this experiment and (3) Wachter's algorithm (Supplementary Fig. \ref{fig_ext:eqi_metrics}). Then, we will be able to propose policies for improving county EQI (Fig. \ref{fig:eqi-over_c}).

The results obtained match those of the experiments using the synthetic benchmark: BayesACE tends to find significantly more actionable and sparser counterfactuals for lower penalties (values of 3 and 5) but for higher ones (values of 10 and 15) FACE with the NF and Wachter's algorithm surpass BayesACE in those two metrics, respectively (Supplementary Fig. \ref{fig_ext:eqi_metrics}). In our experiments, we will focus on a penalty of 5, unless explicitly stated.

\subsubsection{Bayesian network classifier structure}
\begin{figure}[!t]
    \centering
    \includegraphics[width=.85\linewidth]{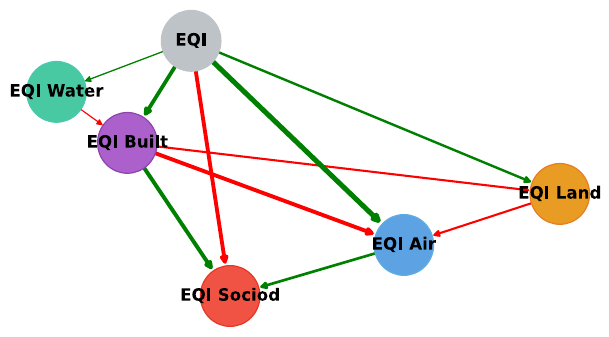}
    \caption{Bayesian network EQI structure. The general EQI has a positive coefficient with the rest of them (green arcs) except for sociodemographic one (red arc), showing how improving the general EQI may worsen the sociodemographic axis. Other notable coefficients include the positive one between the infrastructure and the sociodemographic domain.}
    \label{fig:eqi-exp_graph}
\end{figure}

The five individual quality indices, namely air, water, land, built (also called infrastructure), and sociodemographic, are represented in the data along the joint EQI. The graph can help us understand how these indices interact and are correlated (Fig. \ref{fig:eqi-exp_graph}). 

We also inspected how the variables of each domain relate to its indicator. For the air, land, and water domains, almost all the relations between the quality index and the variables are inversely proportional, since most variables refer to the concentrations of miscellaneous pollutants (Supplementary Fig. \hyperref[fig_ext:structs]{5a-c}). We find a great complexity among the infrastructure, although the variables with higher coefficients are again coherent with the technical report on the dataset \cite{eqi2020}. For example, social-service and civic-related businesses positively affect the infrastructure index, while the commute time of citizens or the proportion of bad roads is negative (Supplementary Fig. \hyperref[fig_ext:structs]{5d}).  In the sociodemographic axis, most relations are consistent with the interaction described in the EQI technical report (Supplementary Fig. \hyperref[fig_ext:structs]{5e}).

\subsubsection{Proposal of policies to improve EQI}
We divide counties based on the EQI into seven categories, from the most developed to the least developed (a lower category means a better quality of life). Our objective is to find in which domain to develop policies and which variables are the most significant to improve the counties, placing them in two categories below. We also find different policies for counties belonging to different categories and to different urbanization levels (rural-urban continuum code, RUCC, where 1 indicates highly urbanized and 4 sparsely populated). 

\begin{figure}[!t]
    \centering
    \subfloat[]{\includegraphics[width=\linewidth]{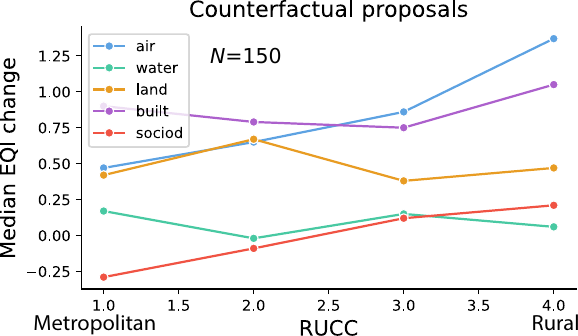}
    \label{fig:eqi-exp_lines_a}}

    \subfloat[]{\includegraphics[width=\linewidth]{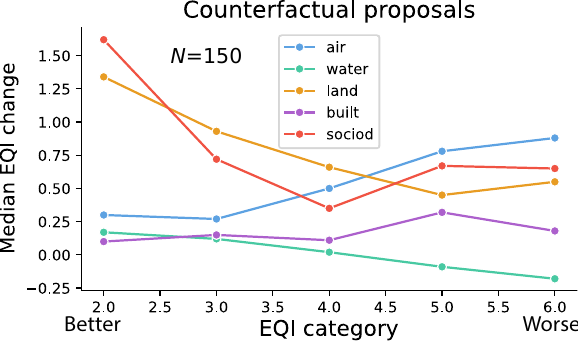}
    \label{fig:eqi-exp_lines_b}}
    
    \caption{Line plots of the BayesACE proposals segregated by RUCC, i.e, urbanisation level (a) and current EQI category (b). For more metropolitan areas, a loss in the sociodemographic index is predicted when improving general EQI, whereas for more rural one this index is improved. A greater overall investment is needed in more rural areas. When analyzing based on the current EQI category, the need for water and land improvements increase for low EQI counties, whereas the need for a sociodemographic and air quality improvement decreases ($N=150$). 
    }
    \label{fig:eqi-exp_lines}
\end{figure}

Proposing different actions depending on whether the county is more or less urbanized is also a sensible choice (Fig. \ref{fig:eqi-exp_lines_a}). In order to improve metropolitan areas, all individual EQIs should be improved, safe for the sociodemographic, where a slight decrease is expected. In contrast, for more rural areas, the sociodemographic index should be improved. A larger improvement in the other domains is also needed, especially in the air domain. This is likely due to the higher presence of heavy industry, explained by the decentralization of manufacturing jobs in previous years~\cite{barkley1995economics}.

When analyzing counterfactuals found for each EQI category, we find that counties with an already good EQI require a higher focus on the sociodemographic domain, while in worse counties a uniform improvement across all indices is needed, with the air quality being the most important (Fig. \ref{fig:eqi-exp_lines_b}).

It is noteworthy that the decrease that occurs in the sociodemographic axis in some scenarios is not always detected by the non-actionable algorithm, potentially harming this axis without realizing it when applying the policies. FACE does not contradict any of the results found with BayesACE, proposing similar counterfactuals but with a greater need of changing the individual indices (higher Euclidean distance). 

We also study which variables specifically change the most for each domain (Supplementary Fig. \ref{fig_ext:vars}). For the air, water, and land domains, BayesACE mainly relies on reducing the quantities of toxic elements present in those domains, such as reducing the tons of isophorone emitted and the presence of cobalt compounds, which are correlated in the structure of the Bayesian network classifier. The case of the sociodemographic index is especially interesting, as that index is reduced in some experiments. The variables that tend to be significant for the index are the percentage of vacant houses, the median value of the home and the percentage of home ownership. All of these variables are related to the access to and quality of the real-state market.  Thus, our algorithm successfully detects that what appears to be a minor decrease in the sociodemographic index can have a noticeable impact on the housing of communities, which is particularly interesting in the context of the current worldwide housing crisis~\cite{ccelik2024cracking,coupe2021global}.

\subsection{Use case: improving the EQI in Los Angeles, New York and a rural area}
To better illustrate our proposal, we focus on the improvement of the well-known counties, Los Angeles and New York, and a vastly more rural one: Conejos, Colorado. All of them are labeled with an initial EQI category of 3 and we plan to improve this category to 1. We will again focus on a penalty value of 5. Actionable recourses (changes to apply) are shown in Supplementary Fig. \ref{fig_ext:eqi_practical}.

The results are consistent with the Supplementary Fig. \hyperref[fig_ext:eqi_metrics]{4a}, with BayesACE surpassing FACE and Wachter's algorithm in almost all metrics for the three analyzed counties. However, for New York County, FACE shows slightly better actionability and sparsity than BayesACE.

It should be noted that, for Los Angeles County, FACE contradicts BayesACE and proposes a policy that slightly increases the sociodemographic domain. Given that BayesACE yields a more numerically actionable result, this indicates that an increase in that index while improving the overall EQI might not be possible. This conclusion is further supported by the fact that increasing the penalty to 10 or 15 results in FACE detecting this decrease in the sociodemographic index.

The need for investment in New York and Los Angeles is different. BayesACE prioritizes air quality for New York, which is logical given that it is an extremely large city. Even if Los Angeles is similarly urbanized, the needs of the county revolve around infrastructure and water quality improvements (depicted in Fig. \ref{fig:eqi-over_c}). The latter is particularly consistent with 70\% of the river flow coming from wastewater~\cite{wolfand2022dilution}. Specifically, BayesACE proposes to reduce the average concentrations of methoxychlor and dinoseb (pesticides), thus suggesting policies focused on local agriculture.

For Conejos, the rural county studied, Wachter's algorithm suggests investing in the sociodemographic index, which is an index unchanged by Bayes and FACE, suggesting its non-actionability. BayesACE focuses their efforts on improving infrastructure, land and water quality in that order, without damaging any of the other indices. With regard to infrastructure, BayesACE's main proposal aims to improve public transport connections and road quality.

\section{Discussion}
Assuming our density estimation and path planning algorithm are fine-tuned, our method is much more expressive and complete than the current state-of-the-art, as the results show that it is able to explore theoretically an infinite number of data points with minimal memory burden. 

DAACE also proves an empirical independence on an abundance of data points, only needing 1 or 2 intermediate points. As such, the question changes from working with a lot of data points to find actionable paths to searching for a small number of quality middle points.

Although the FACE algorithm finds counterfactuals that are connected by high-density regions, it suffers from limitations, such as the overreliance on training endogenous points. The counterfactuals found with DAACE are very competitive against all instances of FACE, surpassing it in almost all metrics in experiments with synthetic data (Figs. \ref{fig:synt-exp} and \ref{fig:synt-exp-l2}).

A more significant result is the fact that BayesACE can find better counterfactuals with a less accurate density estimator in some metrics and scenarios, performing even better than FACE GT while adding enhanced interpretability (Fig. \ref{fig:eqi-exp_graph}). When comparing BayesACE with DAACE, we find ourselves with the well-known interpretability and performance trade-off \cite{gunning2019xai}, with DAACE (using the NF) optimizing better actionability and BayesACE offering the explainability benefits of probabilistic graphical models.

Proposals on actionability also consider classical objectives of counterfactual optimization, such as similarity or likelihood. FACE and DAACE require the user to provide two thresholds, on the likelihood and on the posterior probability of the class given the feature. FACE also includes the connectivity parameter $\epsilon$, a threshold that indicates which data points we should connect to the graph in which FACE seeks counterfactuals, creating a trade-off between actionability and similarity. DAACE however, balances these objectives with the penalty parameter. It is safe to assert that our proposal is more competitive, as DAACE is shown to even surpass Wachter's algorithm for low penalty values, while in contrast FACE is just competitive at best (Figs. \ref{fig:euclidean_line} and \ref{fig:synt-exp-l2}).

Other literature resorts to multi-objective approaches ~\cite{dandl2020multi, dandl2022multi} in order not to neglect other desirable counterfactual properties, with some of them mentioning (but not implementing) actionability as defined by FACE~\cite{rasouli2024care}. Our proposal is flexible enough to add new objectives with minimal implementation changes and parallel computating, which was done in preliminary experiments.

A notable result is that in many experiments with synthetic data (Figs. \ref{fig:synt-exp}, \ref{fig:euclidean_line}, \ref{fig:synt-exp-l2} and Extended Figs. \ref{fig_ext:bayesace_logl},  \ref{fig_ext:bayesace_l2}) the median result for BayesACE is lower (better) than for other algorithms. However, the difference with those algorithms is not significant and in some scenarios they even surpass BayesACE (for instance, Supplementary Fig. \hyperref[fig_ext:bayesace_logl]{2c}). This might indicate that BayesACE is sometimes not able to find (good) counterfactuals given that its estimation of the data density is weaker than for other models in our experiments (Supplementary Fig. \hyperref[fig_ext:model_perf]{1a}). However, when counterfactuals are actually found, they are on average better, hence the lower median. A possible solution would be to use semiparametric Bayesian networks~\cite{atienza2022semiparametric}.

As such, BayesACE might be a better option in scenarios where the data is either jointly Gaussian or faithfully approximated by Gaussians, as it may surpass alternatives (FACE) and the explanation path also tends to be simpler, as only a single (or no) vertex is required.

The results obtained using the EQI dataset are particularly striking. In scenarios like improvement of low EQI counties, traditional non-actionable algorithms (and FACE in some experiments) ignore that a general improvement can result in worsening the specific sociodemographic index. Considering that many of the relevant sociodemographic variables are related to housing access, detecting a possible decrease in this aspect will ensure fairness and equity in policies.


\section{Conclusion}
DAACE and BayesACE are able to successfully find actionable counterfactuals by exploring an estimation of the data distribution rather than directly leveraging on the data. This was done by exploring the log-likelihood function using path planning, a very flexible framework given the multiplicity of existing techniques. 

In our real use case using the EQI dataset, BayesACE performs better than FACE (Supplementary Fig. \hyperref[fig_ext:eqi_metrics]{4a}), even if the latter uses a significantly more accurate NF, yielding simpler counterfactuals. In addition, the structure allows us to visually understand and to help expert validate the relations learned (Fig. \ref{fig:eqi-exp_graph}). The results obtained also match existing literature and studies regarding the importance of housing \cite{ccelik2024cracking, coupe2021global,barkley1995economics} and the state of the environment of specific counties \cite{wolfand2022dilution}.

Despite the advantages of the proposed approach, some limitations must be acknowledged. The quality of the counterfactuals is inherently dependent on the learned data estimator and the penalty parameter can be difficult to control and thus a scan of the penalty values by the user might be necessary.

To further enhance the capabilities of Bayesian networks, a promising direction for future research is to integrate causality into the process. Additionally, we acknowledge the possibility of expanding the framework towards counterfactual inference for other data types, such as images or text.

\section*{Coda availability}
The developed code was formatted as a library, available on GitHub\footnote{\url{https://github.com/Enrique-Val/Bayes-ACE}}. The scripts necessary to run the experiments are also available.

\section*{Acknowledgments}
This work was supported by the Ministry of Education through the University Professor Training (FPU) program fellowship (Enrique Valero-Leal, grant reference FPU21/04812).

This work was also partially supported by the Ministry of Science, Innovation and Universities under Projects AEI/10.13039/501100011033-PID2022-139977NB-I00, TED2021-131310B-I00, and PLEC2023-010252/MIG-20232016. Also, by the Autonomous Region of Madrid under Project ELLIS Unit Madrid and TEC-2024/COM-89.

The authors gratefully acknowledge the Universidad Politécnica de Madrid (www.upm.es) for providing computing resources on Magerit Supercomputer.

\bibliographystyle{unsrt}
\bibliography{references-ijar}

\clearpage
\newpage

\appendix
\section{Supplementary materials}
\setcounter{figure}{0}
\renewcommand\figurename{Supplementary Fig.}
\setcounter{table}{0}
\renewcommand\tablename{Supplementary Table}
\label{sec:supplementary}
The supplementary information is organized as follows. First, we give details on how we created our synthetic datasets by preprocessing and resampling existing ones. Then, we explain how statistical significance was studied for each case, justify the test selection and briefly present critical difference diagrams, our main tool to visualize test results. Finally, we attach additional figures and tables related to the work. These figures are divided into two subsections:
\begin{itemize}
    \item The first subsection contains figures and tables that expand on the results of the main paper or support them visually.
    \item The second subsection consists of figures that contain information about hypothesis test results and tables with the information necessary to reproduce the work. 
\end{itemize}

\subsection{Data preprocessing}
\label{sec_supp:data_prep}
The 15 datasets belong to the numerical classification benchmark of Grinztajn et al.~\cite{grinsztajn2022tree}. We slightly preprocessed the datasets in order to eliminate features that were inconsequential in our experiments. As such, we removed features that met one of the following conditions:
\begin{itemize}
    \item The feature has less than 20 unique values.
    \item More than 90\% of the data values are concentrated in 3 bins, assuming that we discretized the data in 100 bins of equal size.
\end{itemize}

We learn their distribution using kernel density estimation and resampled them. Throughout this work, we work with the resampled data rather than with the original, allowing us to consider the kernel density estimation as our ground-truth distribution. 

For the kernel density estimation we use a Gaussian kernel with a maximum of 10000 instances per class label (selected at random) with a different bandwidth per dataset. We select the kernel bandwidth that minimizes the negative log-likelihood of the data using 10-fold cross-validation, in a range from 0.1 to 1.

Applying the Henze-Zirkler test to all resampled datasets yields a p-value lower than $3\times 10^{-7}$, thus rejecting the hypothesis that the data being distributed as a multivariate normal with a confidence of 5 sigmas to verify with (near) absolute certainty that the data does not follow a multivariate normal distribution. A summary of the datasets is presented in Supplementary Table \ref{tab:datasets}.

\begin{table*}[!h]
    \centering
    \caption[Description of the datasets used for benchmarking.]{Description of the datasets used for benchmarking. The column ``OpenML ID'' refers to the ID in OpenML of the original dataset. We preprocess each dataset, learn its density via kernel density estimation and sample from it, working with the resample rather than with the original one. The number of features and instances shown refers to the total after preprocessing and resampling. For all datasets, we sample as many instances as in the original dataset, but clipping the lower and upper number to 15,000 and 50,000 instances, respectively.}
    \label{tab:datasets}
    \begin{tabular}{lccc}
        \toprule
        \bfseries\makecell{Dataset name} & \bfseries\makecell{OpenML \\ ID} & \bfseries\makecell{Number of \\ features ($n$)} & \bfseries\makecell{Number of \\ instances ($N$)} \\
        \midrule
        credit & 44089 & 4 & 16,714 \\
        california & 44090 & 6 & 20,634 \\
        wine & 44091 & 11 & 15,000 \\
        electricity & 44120 & 6 & 38,474 \\
        covertype & 44121 & 10 & 50,000 \\
        pol & 44122 & 18 & 15,000 \\
        house\_16H & 44123 & 15 & 15,000 \\
        kdd\_ipums\_la\_97-small & 44124 & 8 & 15,000 \\
        MagicTelescope & 44125 & 10  & 15,000 \\
        bank-marketing & 44126 & 7  & 15,000 \\
        phoneme & 44127 & 5  & 15,000 \\
        MiniBooNE & 44128 & 9  & 50,000 \\
        Higgs & 44129 & 24  & 50,000 \\
        eye\_movements & 44130 & 18  & 15,000 \\
        jannis & 44131 & 54  & 50,000 \\
        \bottomrule
    \end{tabular}
\end{table*}

\subsection{Statistical testing of hypotheses and visualization}
\label{sec_supp:hyp_test}
The Friedman test \cite{demvsar2006statistical, friedman1937test} allows us to find statistically significant differences between the performance of multiple algorithms. The Bergmann-Hommel post hoc procedure \cite{bergmann1988improvements} detects more true differences than the standard Friedman test and other post hoc methods by taking into account the family-wise error rate.

Since the total number of p-values $P$ per analysis is 21 (cardinality of the triangular matrix of the 7 algorithms p-values), we present the results using critical difference diagrams~\cite{demvsar2006statistical}. The algorithm performance is ranked from best to worst and a horizontal line between couples of algorithms with non-statistically significant different performance is drawn. We consider a performance difference significant enough if the p-value is less than 0.05, thus rejecting the null hypothesis of equal performance.

The cases where we do not use the Friedman test are typically the ones when we compare only two sets of results and thus the Wilcoxon test is preferred \cite{demvsar2006statistical}. Similarly, we reject the null hypothesis of similar performance of the algorithms when the p-value is smaller than 0.05.

\clearpage
\newpage

\onecolumn

\subsection{Extended Figures and Tables}
\begin{figure}[!h]
    \centering
    \subfloat[]{\includegraphics[width=.95\linewidth]{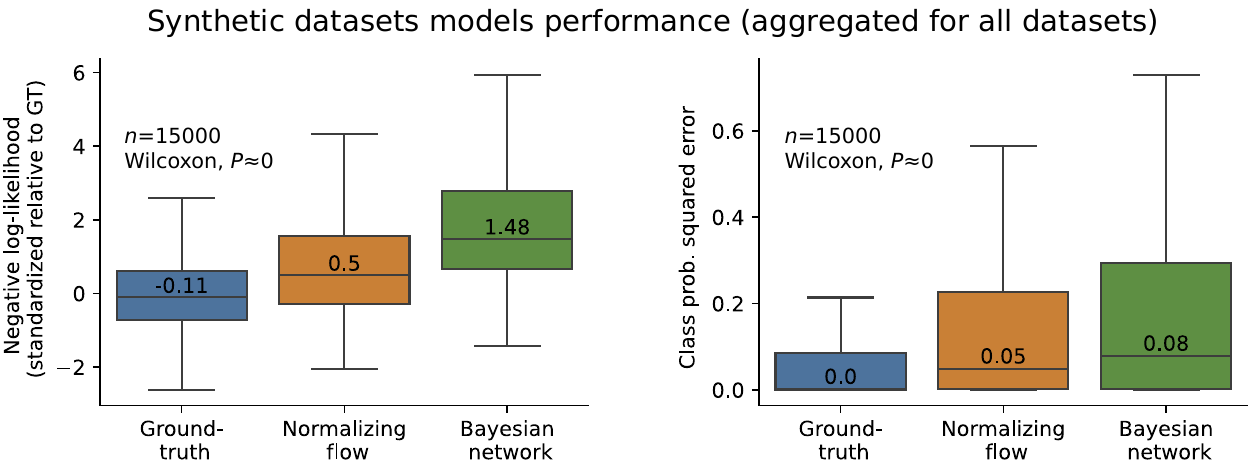}
    \label{fig_ext:model_perf_a}}

    \subfloat[]{\includegraphics[width=.95\linewidth]{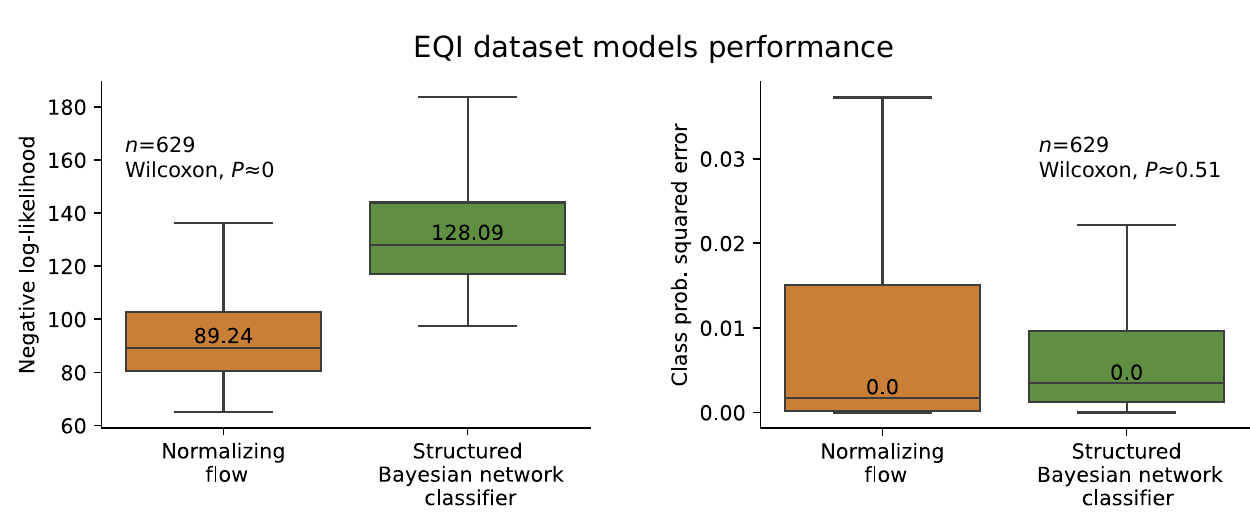}
    \label{fig_ext:model_perf_b}}

    \caption{Model performance. (a) on the left, the negative log-likelihood of the models used standardized with the mean and standard deviation for each of the 15 synthetic datasets generated by sampling from kernel density estimation. On the right, the distribution of the squared error of the class conditional probability given the features. The best performance are achieved, in order, by the ground truth (a kernel density estimator), the NF and the Bayesian network, tested using the Wilcoxon test with a p-value close to 0 ($N=15000$, 1000 samples per dataset). (b) The performance of the NF and the Bayesian network classifier trained on the EQI dataset. Since the data is not synthetic, there is not a ground-truth model.}
    \label{fig_ext:model_perf}
\end{figure}

\newpage
\clearpage

\begin{figure*}[!h]
    \centering
    \subfloat[]{\includegraphics[width=.95\linewidth]{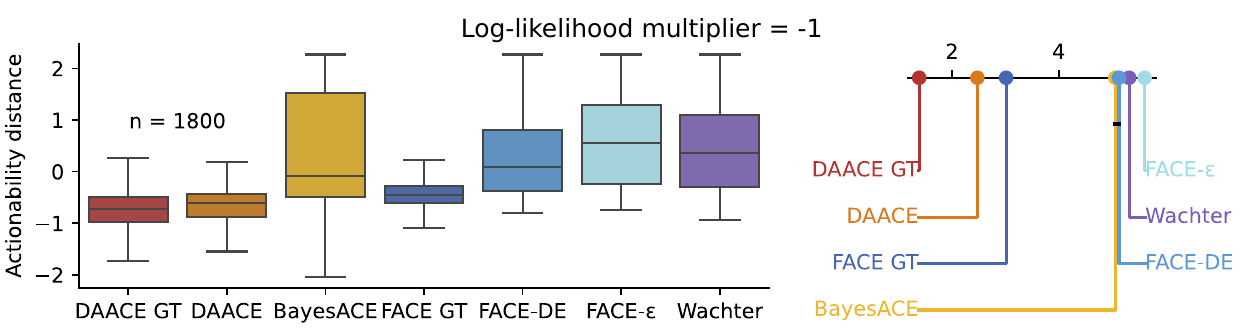}
    }

    \subfloat[]{\includegraphics[width=.95\linewidth]{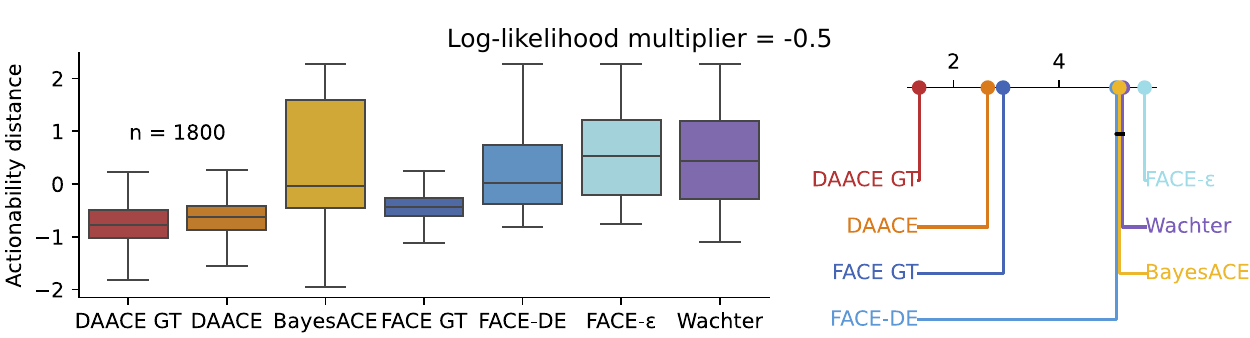}
    }

    \subfloat[]{\includegraphics[width=.95\linewidth]{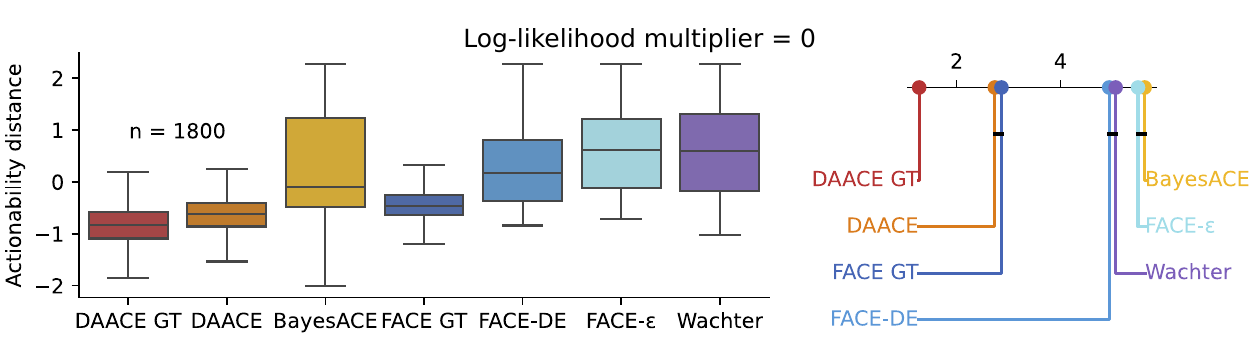}
    }

    \caption{Box plot of actionability distances. (a-c) Results segregating by log-likelihood threshold multiplier. The critical difference diagrams evidence the FACE-DE only surpass BayesACE for a higher log-likelihood threshold. For lower ones, both algorithms are competitive, even when the NF estimates the data density significantly better than the Bayesian network classifier (Supplementary Fig. \ref{fig_ext:model_perf_a}). It is noteworthy that the median distance is always lower for BayesACE (-0.09, -0.04 and -0.09 for the respective log-likelihood multipliers) than for FACE-DE (0.09, 0.02 and 0.17 respectively). This supports our hypothesis (Discussion section) of BayesACE failing to find counterfactuals in some scenarios, although when they are found they are more similar to FACE GT than the ones found with FACE-DE (hence the lower median).}
    \label{fig_ext:bayesace_logl}
\end{figure*}

\newpage
\clearpage

\begin{figure*}[!h]
    \centering
    \subfloat[]{\includegraphics[width=.95\linewidth]{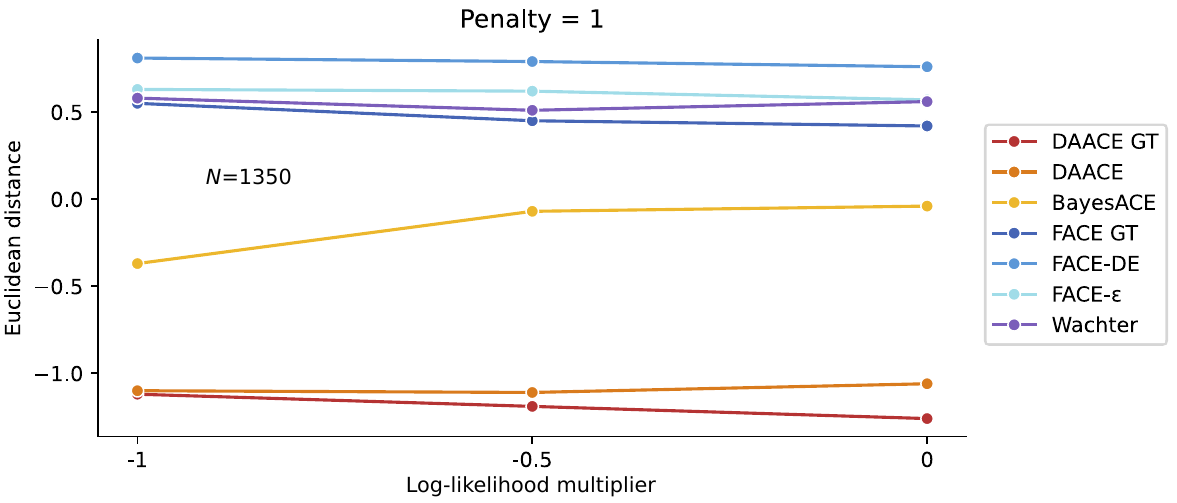}}

    \subfloat[]{\includegraphics[width=.95\linewidth]{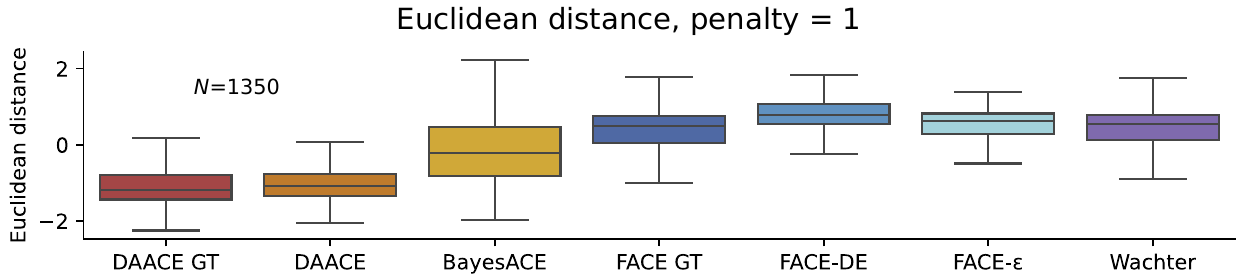}}

    \subfloat[]{\includegraphics[width=.95\linewidth]{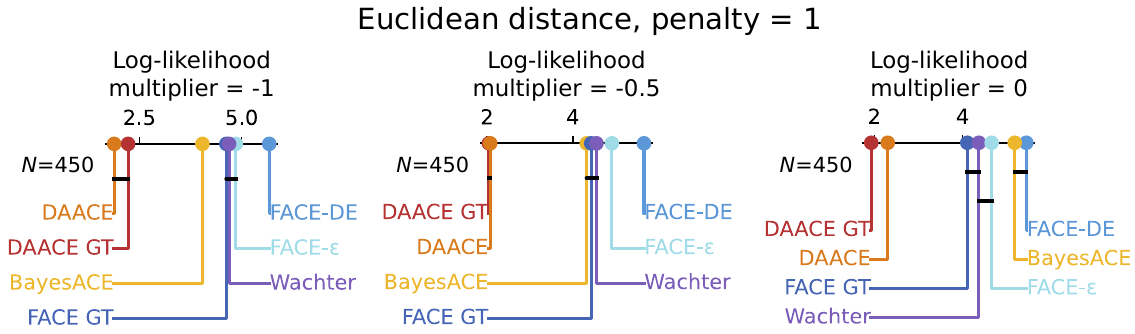}}

    \caption{Additional Euclidean distance results for penalty=1. (a) Line plot representing the Euclidean distance per log-likelihood multiplier value: the median results are fairly similar for all values ($N=1350$). (b) Box plot for the Euclidean distance: the median distances of our proposals, DAACE (GT) and BayesACE, are lower than for the rest of algorithms (BayesACE vs FACE-DE median, $N=1350$). (c) Critical difference diagram segregating by log-likelihood multiplier: BayesACE significantly surpasses the current state-of-the-art proposals in terms of Euclidean distance for low log-likelihoods thresholds. Even when BayesACE becomes statistically worst than its competitors, the median Euclidean distance is still lower. ($N=450$).}
    \label{fig_ext:bayesace_l2}
\end{figure*}

\newpage
\clearpage

\begin{figure*}[!h]
    \centering
    \subfloat[]{\includegraphics[width=.45\linewidth]{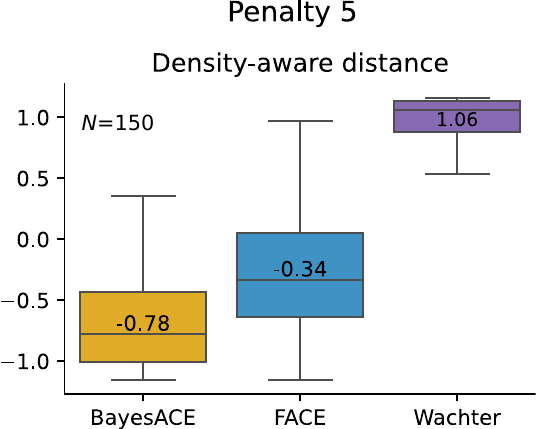}}
    \hfil
    \subfloat[]{\includegraphics[width=.45\linewidth]{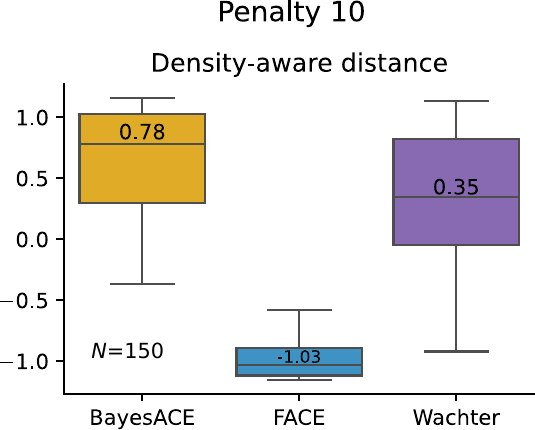}}

    \subfloat[]{\includegraphics[width=.45\linewidth]{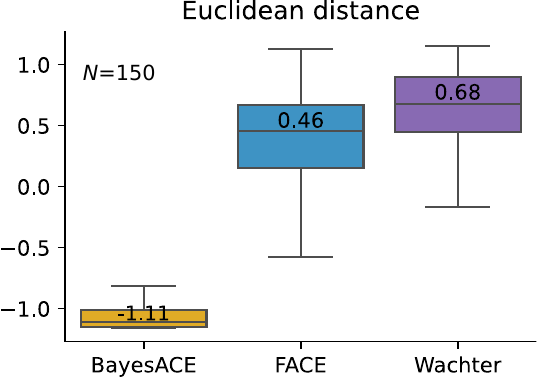}}
    \hfil
    \subfloat[]{\includegraphics[width=.45\linewidth]{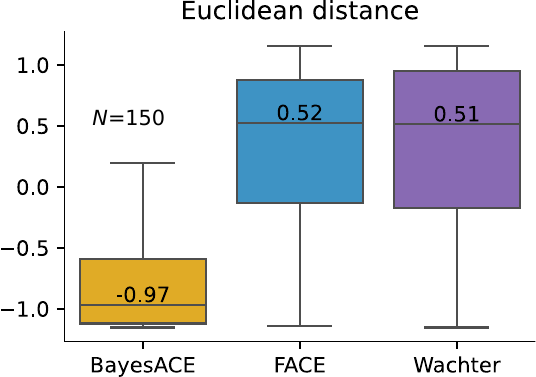}}

    \subfloat[]{\includegraphics[width=.45\linewidth]{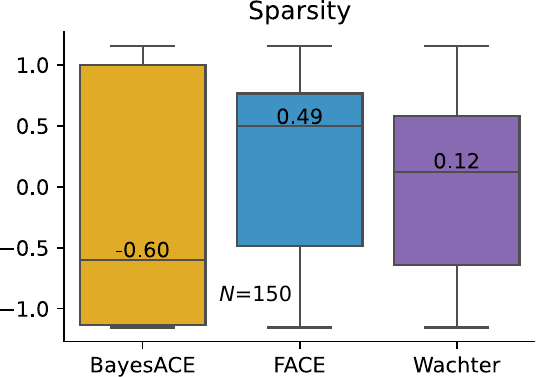}}
    \hfil
    \subfloat[]{\includegraphics[width=.45\linewidth]{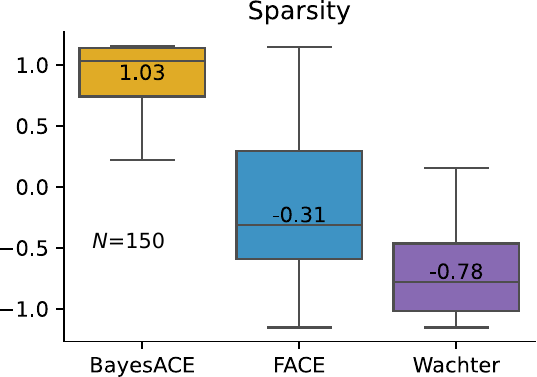}}

    \caption{Performance of the algorithms trained on the EQI dataset with penalty values 5 and 10, left and right columns respectively. (a-b) Comparison of actionability (density-aware distance). With a higher penalty, the density estimation of the CLGN (BayesACE) begins to short, as it can only represent linear relations between variables, Performance with penalty=5: BayesACE performs better than in all the metrics than FACE (using FACE-DE implementation with the NF) and the Wacther's algorithm ($N=150$).
    (c-d) Sparsity comparison. For both penalty parameter values, BayesACE outperforms FACE and the Wachter's algorithm.
    (e-f) Sparsity comparison (L0 distance). Similarly as with the actionability, the performance of BayesACE degenerates with higher penalty values ($N=150$ for all experiments).}
    \label{fig_ext:eqi_metrics}
\end{figure*}

\newpage
\clearpage

\begin{figure*}[!h]
    \centering
    \subfloat[]{\includegraphics[width=.45\linewidth]{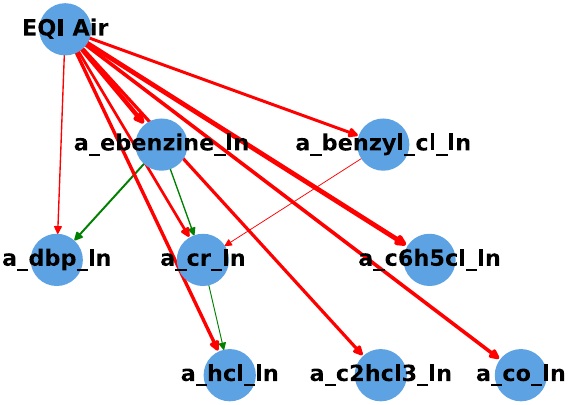}}
    \hfil
    \subfloat[]{\includegraphics[width=.45\linewidth]{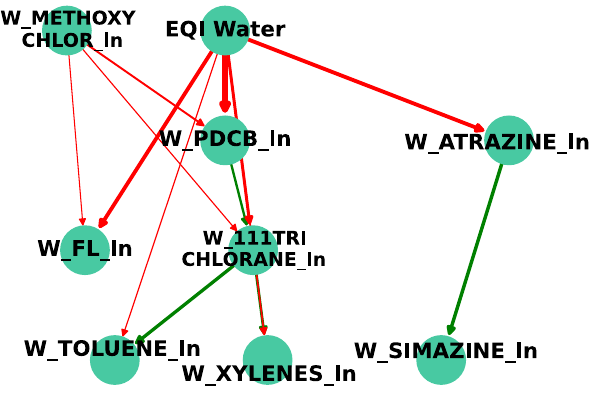}}

    \subfloat[]{\includegraphics[width=.45\linewidth]{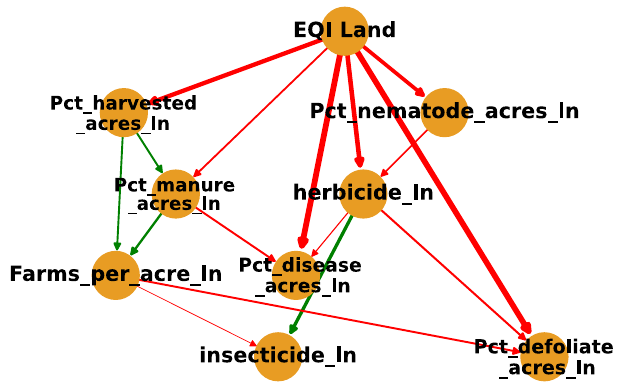}}
    \hfil
    \subfloat[]{\includegraphics[width=.45\linewidth]{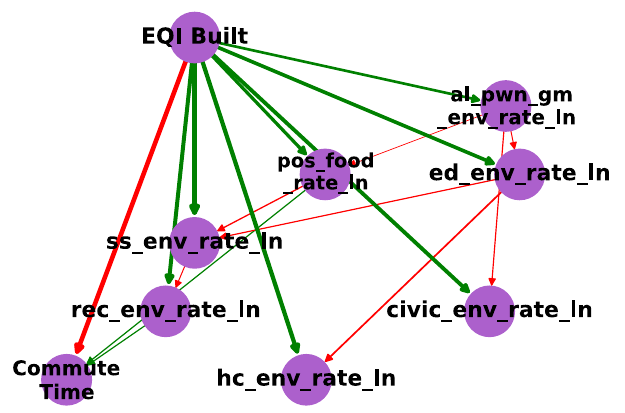}}

    \subfloat[]{\includegraphics[width=.45\linewidth]{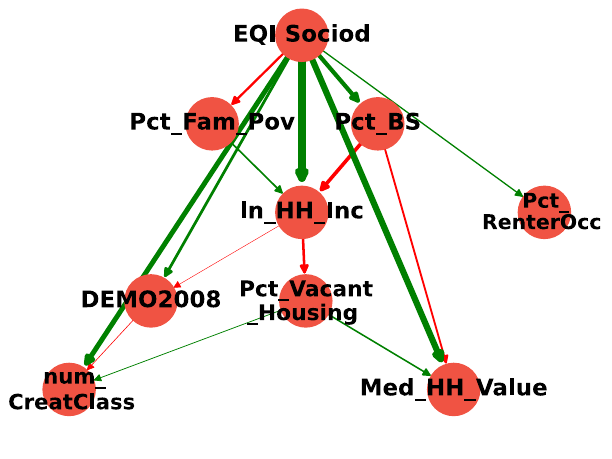}}

    \caption{Bayesian network structure, separated by domains. Since showing all variables/nodes per substructure is unfeasible due to its high number, we show only the 8 variables with higher correlation with the domain-specific index. Green arcs indicate relations with positive coefficients, whereas red is used for negative coefficients. (a-c) Air, water and land domain respectively. Almost all the relations among the EQI and the variables are negative, since they mostly refer to pollutants or pollutants yearly emissions that in higher concentrations affect negatively the environment. (d-e) Built (also referred as infrastructure) and sociodemographic and  domains, where relations with the domain-specific EQI are more diverse.}
    \label{fig_ext:structs}
\end{figure*}

\newpage
\clearpage

\begin{figure*}[!h]
    \centering
    \subfloat[]{\includegraphics[width=.9\linewidth]{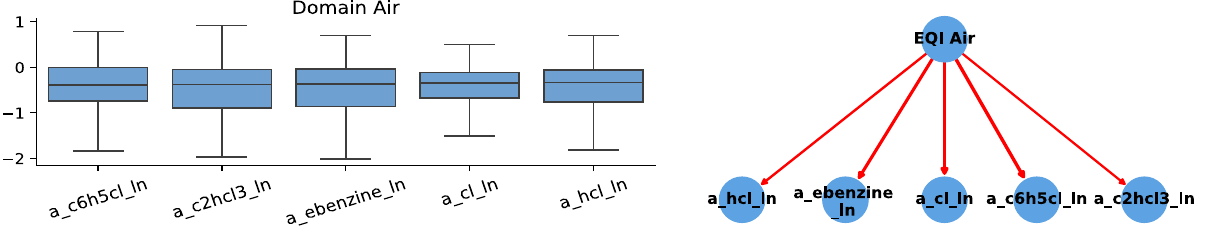}}

    \subfloat[]{\includegraphics[width=.9\linewidth]{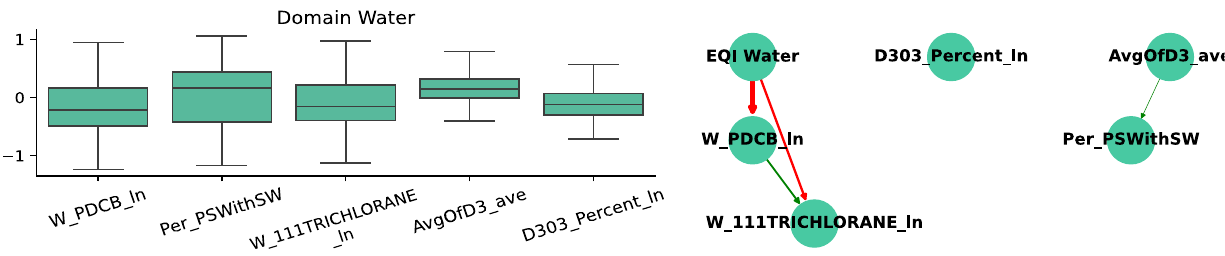}}

    \subfloat[]{\includegraphics[width=.9\linewidth]{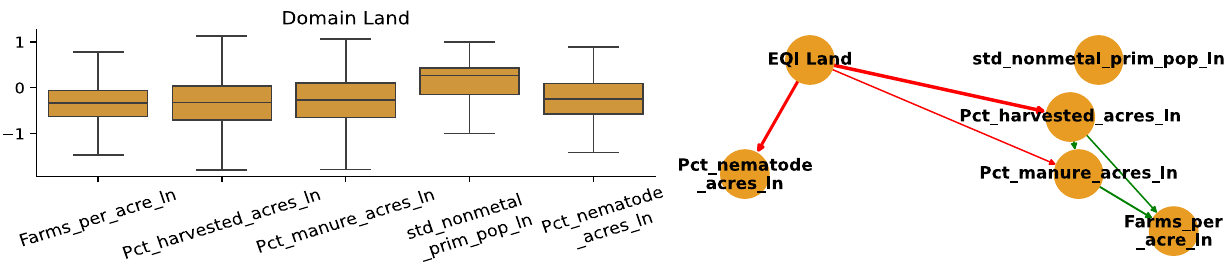}}

    \subfloat[]{\includegraphics[width=.9\linewidth]{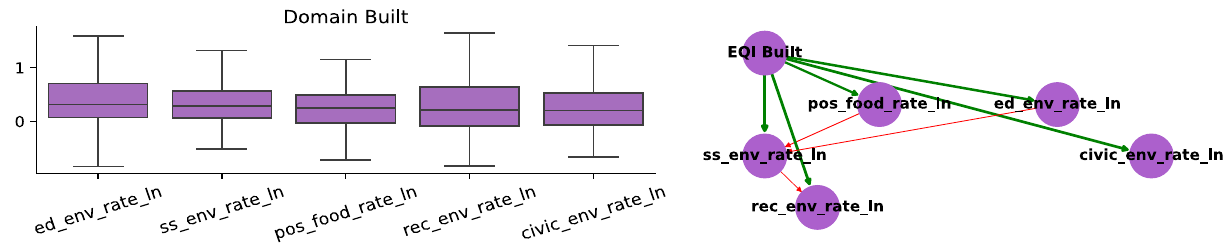}}

    \subfloat[]{\includegraphics[width=.9\linewidth]{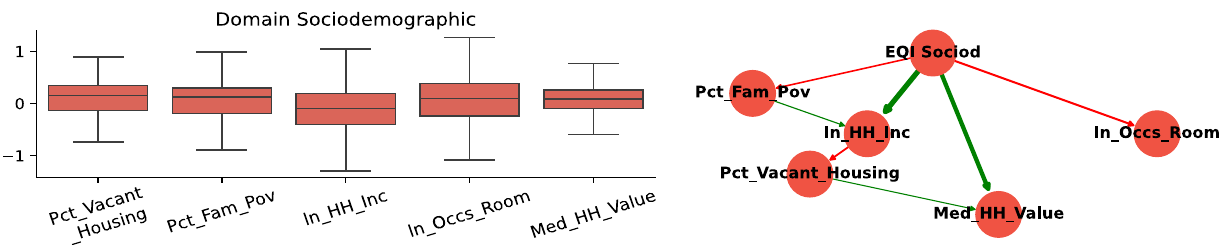}}

    \caption{Most changed variables per domain. On the left, the box plot of the 5 most changed variables per domain applying the BayesACE algorithm with a penalty of 5. On the right, their respective Bayesian network structure showing how they are related. Since graphs are incomplete, some variables may lack a connection with the EQI, which does not occur when plotting the entire structure.}
    \label{fig_ext:vars}
\end{figure*}

\newpage
\clearpage

\begin{figure*}[!h]
    \centering
    \subfloat[]{\includegraphics[width=.85\linewidth]{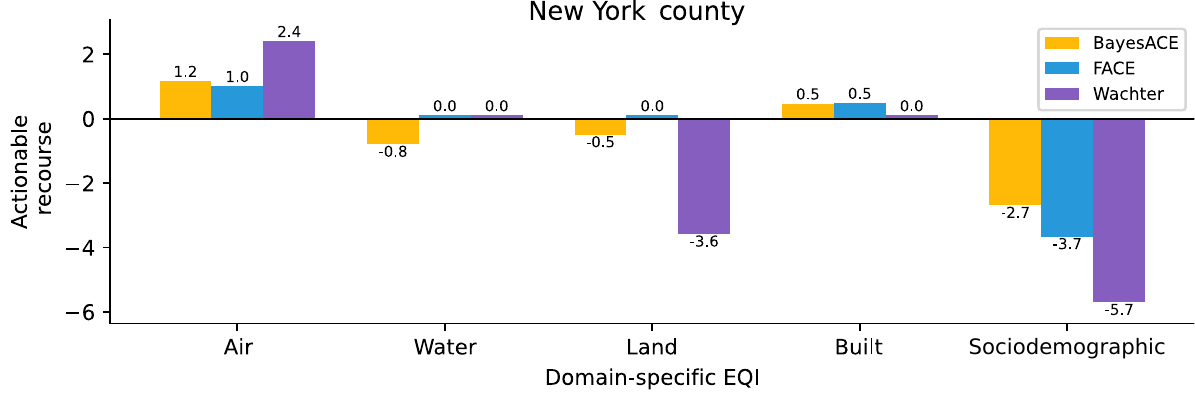}
    \label{fig_ext:eqi_practical_ny}}

    \subfloat[]{\includegraphics[width=.85\linewidth]{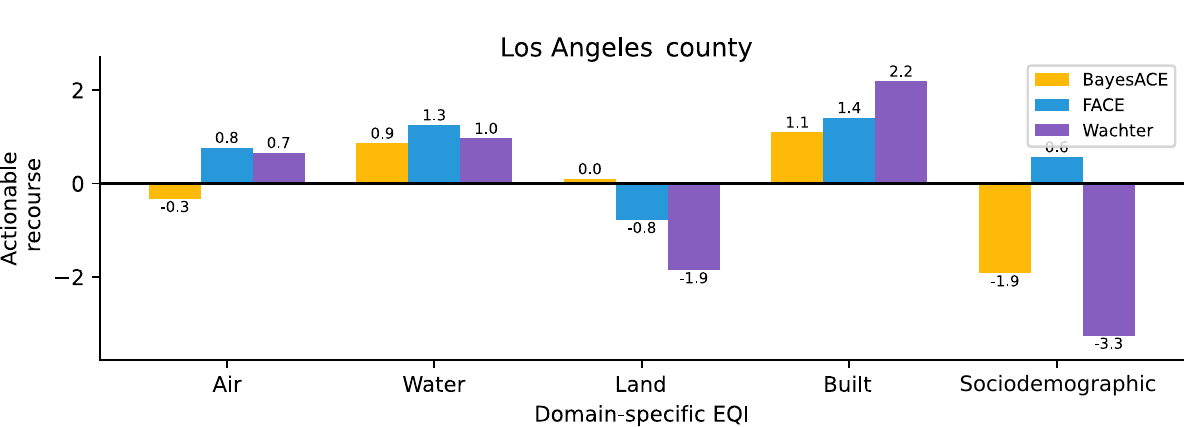}
    \label{fig_ext:eqi_practical_ca}}

    \subfloat[]{\includegraphics[width=.85\linewidth]{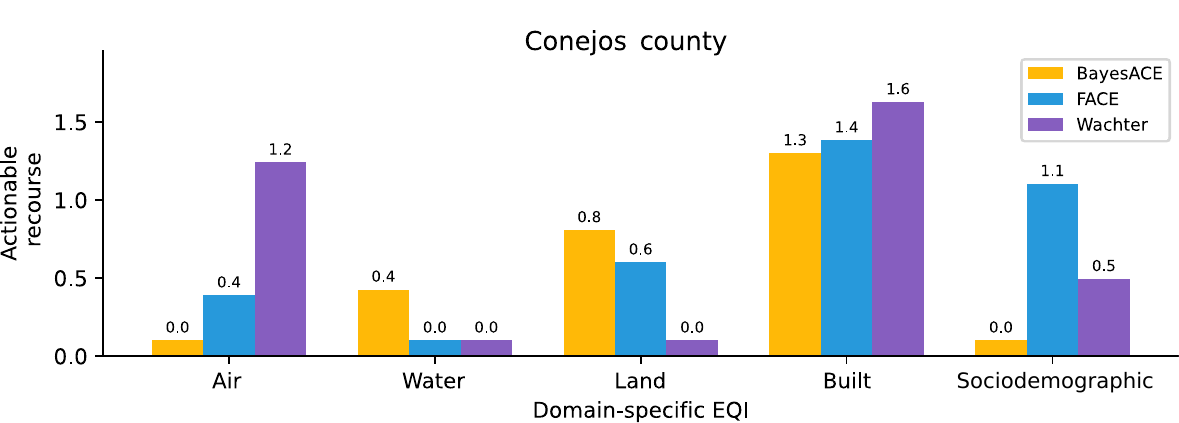}
    \label{fig_ext:eqi_practical_rur}}

    \caption{EQI actionable recourse for three USA counties using three algorithms. (a) New York county: For this example, all three algorithms agree that the air is the main domain-specific index to improve. However, Wachter's algorithm predicts a massive decrease in the land and sociodemographic domain (-3.6 and -5.7), with BayesACE and FACE only agreeing in the decrease in sociodemographic quality decrease, but in a considerable smaller amount (-2.7 and -3.7, respectively). (b) Los Angeles county: Water and infrastructure are the main indices to improve. BayesACE predicts a decrease in the sociodemographic index of 1.9, whereas FACE proposes an increase of 0.6. Given that BayesACE surpasses FACE in all metrics for this example, it is safe to assume that the proposal of FACE might be unrealistic. (c) Conejos county: It is visible how in the rural world, even if its EQI is the same as New York and Los Angeles, a bigger uniform investment across all indices is needed. No index is actually damaged during this increase. Again, BayesACE proposes easier policies (less changes) while still being more actionable, with a mean EQI change of 2.59 against 3.67 and 4.5 for FACE and Wacther's algorithm, respectively.}
    \label{fig_ext:eqi_practical}
\end{figure*}

\newpage
\clearpage

\onecolumn

\subsection{Hypothesis testing and reproducibility}
\begin{figure*}[!h]
    \centering
    \subfloat[]{\includegraphics[width=0.9\linewidth]{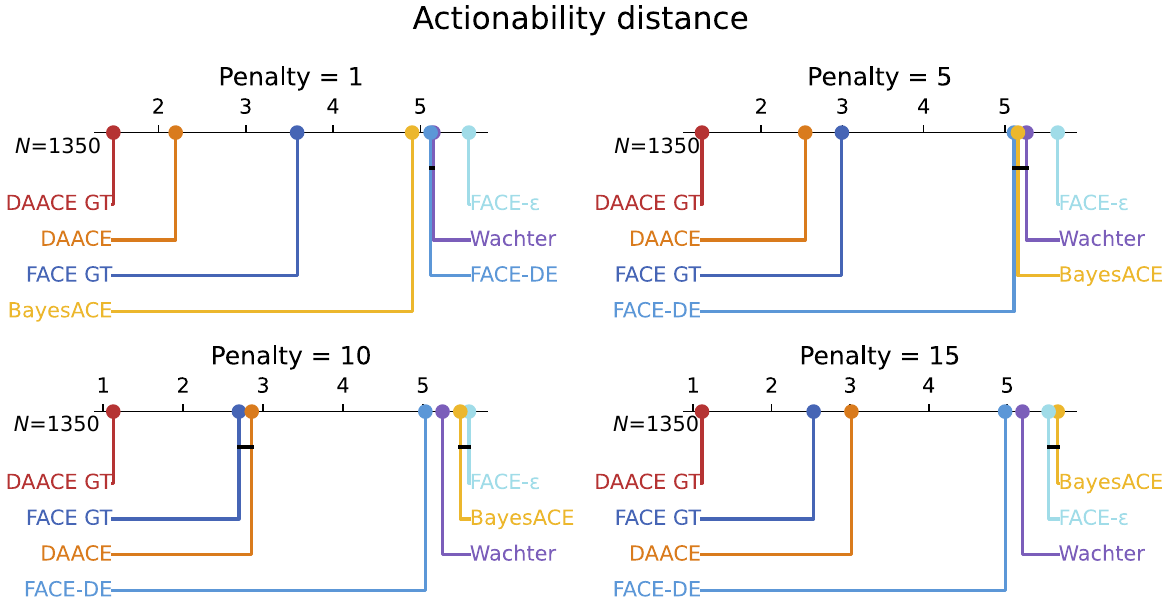}}

    \subfloat[]{\includegraphics[width=0.9\linewidth]{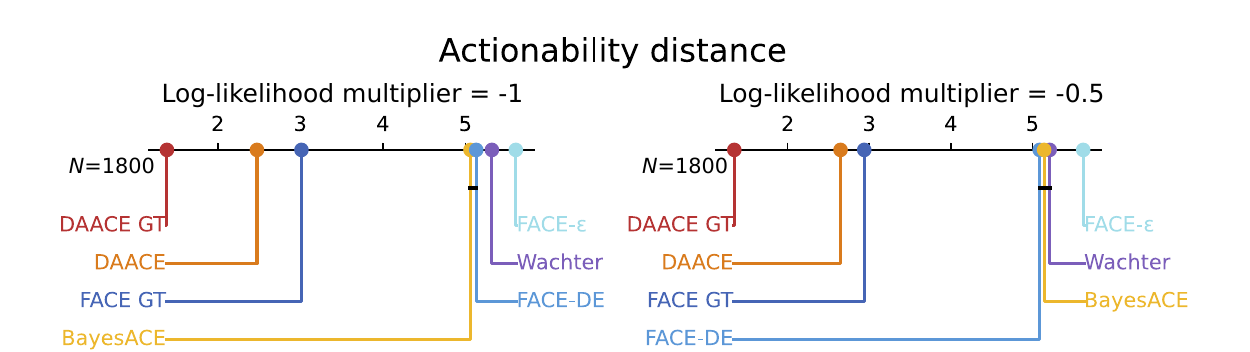}}

    \subfloat[]{\includegraphics[width=0.45\linewidth]{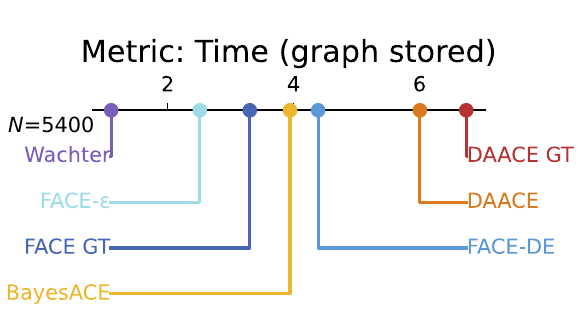}}
    \hfil
    \subfloat[]{\includegraphics[width=0.45\linewidth]{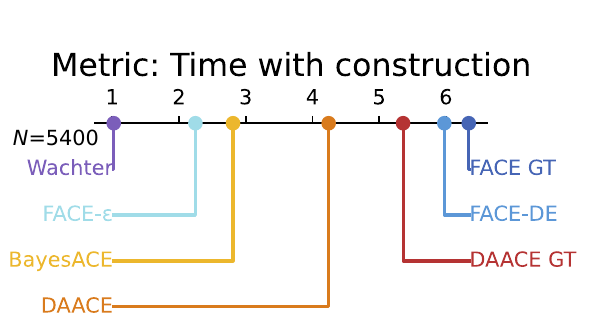}}

    \caption{Critical difference diagrams supporting the results of Fig. \ref{fig:synt-exp}. (a) Actionability distance segregating by penalty ($N=1350$): while DAACE and BayesACE performance drops slightly for higher penalties, the upper bound comparison (DAACE GT vs FACE GT) is always won by DAACE. (b) Actionability distance segregating by log-likelihood threshold multipliers of -1 and -0.5 ($N=1800$): DAACE (GT) is the best model and BayesACE is competitive with FACE-DE in both cases. (c-d) Running time with and without storing the FACE graph in memory ($N=5400$). }
    \label{fig_supp:synth_res}
\end{figure*}

\newpage
\clearpage

\begin{figure*}[!h]
    \centering
    \subfloat[]{\includegraphics[width=.9\linewidth]{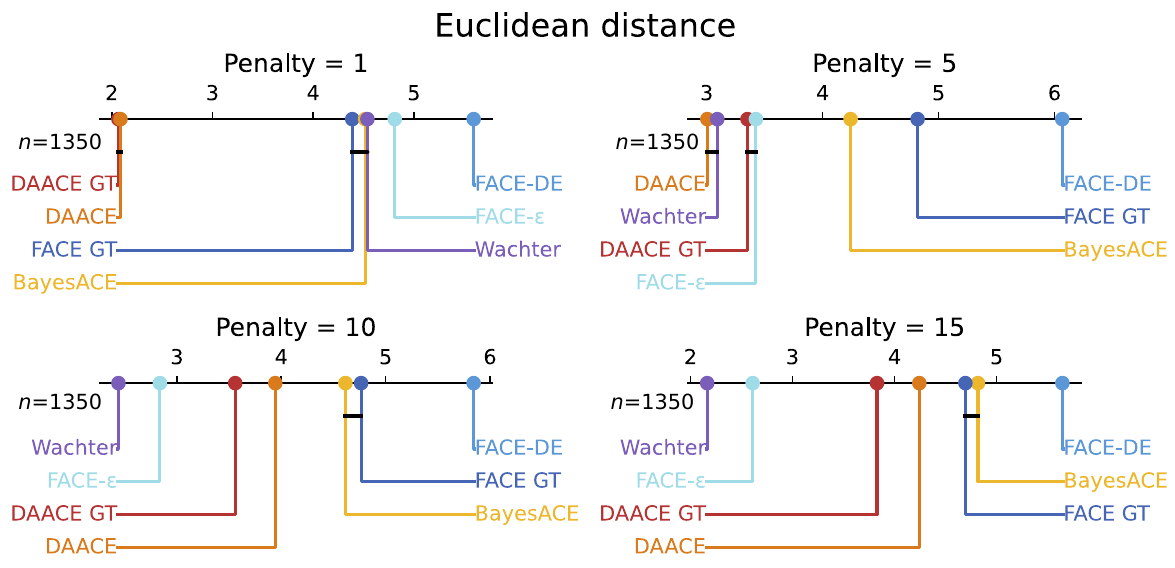}}

    \subfloat[]{\includegraphics[width=.9\linewidth]{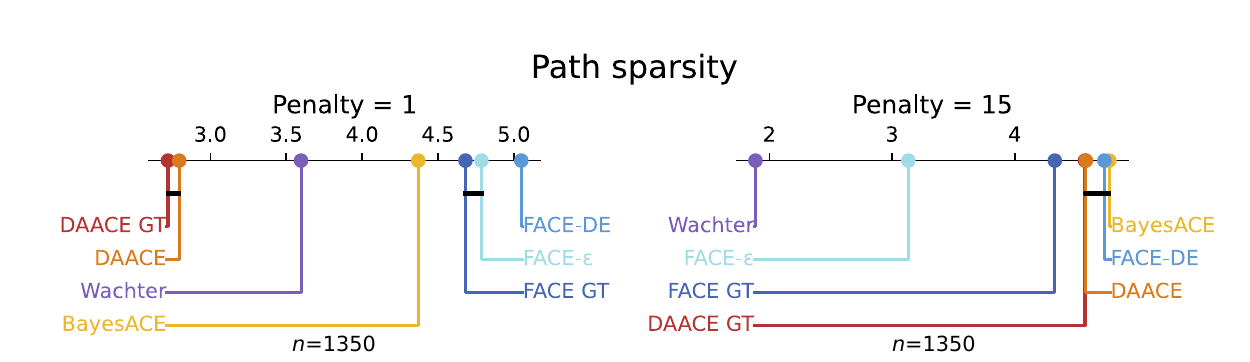}}

    \subfloat[]{\includegraphics[width=.45\linewidth]{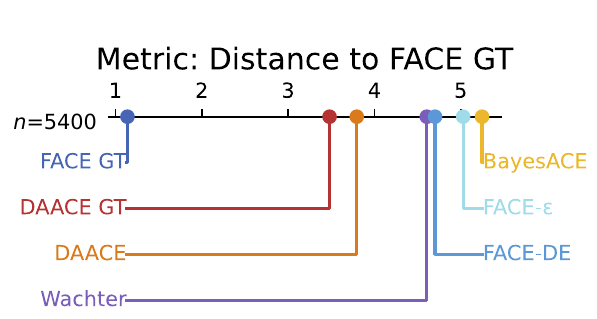}}

    \caption{Critical difference diagrams supporting the results of Figs. \ref{fig:euclidean_line} and \ref{fig:synt-exp-l2}. (a) Euclidean distance segregating by penalty ($N=1350$): DAACE is able to compete or surpass the Wachter's algorithm for lower penalties. BayesACE also proves to find significantly closer counterfactuals than FACE-DE. (b) Path sparsity segregating by penalty (1 and 15 shown, $N=1350$): similarly, our algorithms (DAACE and BayesACE) rank relatively better for lower penalties. (c) Distance (Euclidean) to the counterfactual found by FACE GT ($N=5400$): DAACE ranks better than the FACE specific implementations (FACE-DE and FACE-$\epsilon$). Even if BayesACE ranks worse, the median distance is lower (refer to Fig. \ref{fig:euclidean_line}).}
    \label{fig_supp:synt-l2}
\end{figure*}

\newpage
\clearpage

\begin{table*}[!h]
\centering
    \caption{Bayesian optimization parameter space for the NF. Note that ``Hidden units'' does not refer to the total of neurons per layer, but to a multiplier, such as the total number of neurons is computed as the product of the multiplier and the number of features in the dataset. The upper limit of the standard deviation of the Gaussian noise for the EQI dataset was increased to 0.5.}
    \label{tab:bayes_opt}
    \begin{tabular}{lccc}
    \toprule
    \bfseries\makecell{Parameter} & \bfseries\makecell{Lower \\ bound} & \bfseries\makecell{Upper \\ bound} & \bfseries\makecell{Prior \\ distribution} \\
    \midrule
    Learning rate & $10^{-4}$ & $5 \times 10^{-3}$ & Uniform \\
    Weight decay & $10^{-4}$ & $10^{-2}$ & Uniform \\
    Hidden units  & 2 & 10 & Uniform (Integer) \\
    Layers & 1 & 5 & Uniform (Integer) \\
    Flows & 1 & 10 & Uniform (Integer) \\
    Gaussian noise $\sigma$ & 0.01 & 0.3 & Log-Uniform \\
    \bottomrule
    \end{tabular}
\end{table*}

\begin{table*}[!h]
    \caption{Fine tuned hyperparameters for the NFs. ``Hidden units'' does not refer to the total of neurons per layer, but to a multiplier, such as the total number of neurons is computed as the product of the multiplier and the number of features in the dataset.}
    \label{tab:params_tuned}
    \centering
    \begin{tabular}{lcccccc}
    \toprule
     \bfseries\makecell{Dataset  name} & Learning rate & Weight decay & \makecell{Hidden \\ units} & Layers & Flows & \makecell{Gaussian \\ noise $\sigma$} \\
    \midrule
    credit & 0.00211 & 0.00502 & 6 & 3 & 7 & 0.08688 \\
    california & 0.00079 & 0.00871 & 6 & 4 & 6 & 0.10064 \\
    wine & 0.00221 & 0.00162 & 10 & 5 & 10 & 0.30000 \\
    electricity & 0.00300 & 0.00846 & 9 & 4 & 7 & 0.03696 \\
    covertype & 0.00179 & 0.00800 & 10 & 3 & 7 & 0.14033 \\
    pol & 0.00273 & 0.00333 & 8 & 2 & 7 & 0.19414 \\
    house\_16H & 0.00168 & 0.00754 & 5 & 4 & 10 & 0.17375 \\
     kdd\_ipums\_la\_97-small & 0.00321 & 0.00995 & 7 & 3 & 5 & 0.08337 \\
    MagicTelescope & 0.00089 & 0.00883 & 5 & 4 & 10 & 0.16310 \\
    bank-marketing & 0.00079 & 0.00871 & 6 & 4 & 6 & 0.10064 \\
    phoneme & 0.00228 & 0.00855 & 8 & 4 & 10 & 0.04628 \\
    MiniBooNE & 0.00168 & 0.00989 & 9 & 4 & 9 & 0.10549 \\
    Higgs & 0.00119 & 0.00393 & 9 & 3 & 7 & 0.21522 \\
    eye\_movements & 0.00019 & 0.00361 & 9 & 3 & 7 & 0.26408 \\
    jannis & 0.00119 & 0.00393 & 9 & 3 & 7 & 0.21522 \\
    \midrule
    EQI dataset & 0.00017 &  0.05026 & 3 & 5 & 9 & 0.39740 \\
    \bottomrule
\end{tabular}
    
\end{table*}

\end{document}